\documentclass[journal=dap]{CUP-JNL-DTM}%

\usepackage{graphicx}
\usepackage{multicol,multirow}
\usepackage{amsmath,amssymb,amsfonts}
\usepackage{mathrsfs}
\usepackage{amsthm}
\usepackage{rotating}
\usepackage{appendix}
\usepackage{ifpdf}
\usepackage[T1]{fontenc}
\usepackage{newtxtext}
\usepackage{newtxmath}
\usepackage{textcomp}
\usepackage{xcolor}
\usepackage{lipsum}
\usepackage[colorlinks,allcolors=blue]{hyperref}

\usepackage{csquotes}
\usepackage{array} 
\usepackage{tabularx}
\usepackage{adjustbox} 
\usepackage{caption}
\usepackage{tikz}
\tikzstyle{na} = [baseline=-.5ex]
\usetikzlibrary{arrows}
\makeatletter
\newcommand\@erelb@r[1]{%
  \mathrel{\tikz[baseline=-.5ex]\draw[#1] (0,0)--(0.3,0);}
}
\usepackage{fancyhdr}
\renewcommand\headrulewidth{0pt}

\theoremstyle{definition}

\numberwithin{equation}{section}

\jname{}
\articletype{}
\jyear{}
\begin{document}

\begin{Frontmatter}

\title[Article Title]{Detection Avoidance Techniques for Large Language Models}

% There is no need to include ORCID IDs in your .pdf; this information is captured by the submission portal when a manuscript is submitted. 
\author[1]{Sinclair Schneider}
\author[1]{Florian Steuber}
\author[1]{João A. G. Schneider}
\author[1]{Gabi Dreo Rodosek}

\address[1]{\orgdiv{Research Institute CODE}, \orgname{Bundeswehr University Munich}, \orgaddress{\city{Munich}, \postcode{81739}, \state{Bavaria},  \country{Germany}}. \email{Sinclair.Schneider@unibw.de}}

\authormark{Sinclair Schneider et al.}


\abstract{The increasing popularity of large language models has not only led to widespread use but has also brought various risks, including the potential for systematically spreading fake news. Consequently, the development of classification systems such as DetectGPT has become vital. These detectors are vulnerable to evasion techniques, as demonstrated in an experimental series: Systematic changes of the generative models’ \textit{temperature} proofed \textit{shallow learning--detectors} to be the least reliable (\hyperref[sec:exp1]{Exp. 1}). 
Fine-tuning the generative model via \textit{reinforcement learning} circumvented \textit{BERT-based--detectors} (\hyperref[sec:exp2]{Exp. 2}). 
Finally, \textit{rephrasing} led to a >90\% evasion of \textit{zero-shot--detectors} like DetectGPT, although texts stayed highly similar to the original (\hyperref[sec:exp3]{Exp. 3}). 
A comparison with existing work highlights the better performance of the presented methods. Possible implications for society and further research are discussed.}

\end{Frontmatter}

\section*{Policy Significance Statement}
 Large language models produce texts that appear indistinguishable from human ones, which is why research focuses on machine learning based detectors. 
This paper demonstrates how various state-of-the-art detectors can be tricked using different techniques. Specifically, text-generating models are modified in such a way they a) no longer use the most likely words (parameter \textit{temperature}), b) are penalized for certain conspicuous content (\textit{reinforcement learning}), or c) rephrase sentences so slightly that they remain the same in terms of content but can no longer be recognized as machine-generated (\textit{paraphrasing}).
In short, detectors can easily be bypassed. The implications for society and research are discussed.
Further research is needed to investigate implications, such as the influence on opinion or fake news in social media.

% \localtableofcontents

\thispagestyle{fancy}
\fancyfoot[C]{\textit{Published in Data \& Policy, vol. 7, p. e29, 2025. doi:10.1017/dap.2025.6}}
\renewcommand\headrulewidth{0pt}
\fancyhead{}

\section{Introduction}
As large language models (LLMs) continue to evolve, the necessity for precise differentiation between human outputs and those produced by LLMs is becoming increasingly critical. Recent developments in LLMs have significantly improved, particularly in complex reasoning tasks, such as mathematical problem-solving. These advancements are rapidly closing the performance gap between LLMs and human capabilities, which had previously been a major limitation of such models. Due to this close-up, the difficulty of reliably detecting LLMs has further increased.

Various concepts helped LLMs to catch up to human-like reasoning capabilities. For example, the \textit{Quiet-STaR} approach \parencite{Quiet-STaR} reinforces intermediate beliefs generated by the model before providing a final answer, improving reasoning accuracy. The \textit{Agent Q} framework \parencite{AgentQ} combines \textit{Monte Carlo Tree Search} (MCTS) and \textit{Direct Preference Optimization} (DPO) to teach LLMs to perform complex tasks, such as navigating an online store. Other work, including \textit{Let’s Verify Step by Step} \parencite{LetsVerifyStepbyStep}, enhances LLM reasoning performance by breaking tasks into discrete steps and providing feedback for each step. Likewise, \textit{Verification for Self-Taught Reasoners} \parencite[V-STaR by][]{vstar} refers to a concept in which the LLM generates multiple solutions for a task, learning from correct answers. A verifier model learns from correct and incorrect responses, improving the LLM's reasoning ability, particularly in coding and mathematical tasks.

In light of these rapid advancements, a growing concern is that distinguishing human- from LLM-generated text might become even more challenging.
Models such as DetectGPT \parencite{mitchell_detectgpt_2023} and datasets like the \textit{Human ChatGPT Comparison Corpus} \parencite[HC3, ][]{ChatGPTvsHuman} aim to address this challenge. However, further investigation is required to assess the reliability of these detection models and explore whether they can be circumvented with reasonable effort.
Like cryptography, every detection method, as shown in the current study, is at risk of being attacked and eventually circumvented. 
Such insights lead to the need for additional and robust methods, such as watermarks, to help clarify the origin of published texts.  

This study presents an experimental series to evaluate the reliability of LLM detection models and explore potential methods for bypassing them. In the first experiment (cf. Sec.~\ref{sec:exp1}), shallow learning classifiers are evaluated based on a Bag-of-Words (BoW) approach combined with a Naive Bayes classifier. Although this method is not state-of-the-art, it serves as a benchmark to showcase the influence of hyperparameters, including temperature, sampling method, and model size on classification performance. Additionally, the classifiers' performance is compared to human judgment. Human classification often focuses on identifying unlikely words, while machine models rely on statistical patterns \parencite{DBLP:conf/acl/IppolitoDCE20}. According to \textcite{DBLP:conf/acl/GehrmannSR19}, human judgment achieves an accuracy of only 54\% without automated assistance, improving up to 72\% with supporting tools. Other studies similarly report a 50\% success rate in detecting GPT-3-generated text \parencite{DBLP:conf/acl/ClarkASHGS20}. Using top-p sampling at 1.0, the Naive Bayes classifier's detection rate was reduced below 60\%, indicating that simple models may be bypassed without special techniques.

In the second experiment (cf. Sec.~\ref{sec:exp2}), the combination of shallow feature categories and shallow classifiers is replaced with a BERT-based classifier and corresponding transformer-based embeddings, yielding a significantly improved accuracy exceeding 90\%. This forms the basis for the initial bypassing approach. In contrast to the one described by \textcite{krishna_paraphrasing_2024}, paraphrasing is not used at this stage. Instead, \textit{reinforcement learning} (RL) is employed for model training to preserve the generative model from detection. This approach builds upon the methodology outlined by \textcite{DBLP:journals/corr/abs-1909-08593-neu}, which originally fine-tuned LLMs using human feedback. Special constraints are incorporated into the reward function to prevent the model from learning trivial bypassing strategies, such as adding special characters or introducing artifacts. Depending on the LLM's size, the detection rate reduces from over 90\% to below 17\% following the RL training. This demonstrates that once the classifier is known and accessible, the generative model can be adapted to evade it.

In the third experiment (cf. Sec.~\ref{sec:exp3}), a paraphrasing model is applied, similar to the approach introduced by \textcite{krishna_paraphrasing_2024}. While \citeauthor{krishna_paraphrasing_2024} have focused on general-purpose paraphrasing, the here presented model is tailored to hide the generative model from DetectGPT specifically. 
In this regard, a new dataset is vital. Here, the original LLM output is further paraphrased multiple times. This procedure allows us to select the version least likely to be classified as LLM-generated. Inspired by \textcite{mitchell_detectgpt_2023}, each single paraphrasing iteration altered approximately 15\% of the sentence. Based on the newly created dataset, a paraphrasing model is trained to hide the original language model. By applying this paraphrasing model to the output of the Qwen1.5-4B-Chat model \parencite{qwen_2023}, the detection rate was iteratively reduced from  88.6\% to 8.7\%. A comparative analysis uses the DIPPER model from \textcite{krishna_paraphrasing_2024}. Hereby, the presented approach preserves a higher degree of linguistic similarity to the original text, even after multiple paraphrasing iterations.

In conclusion, this study demonstrates that RL and paraphrasing techniques can effectively bypass LLM detection classifiers. These results suggest that a classifier can easily be bypassed with sufficient knowledge. This can be achieved by fine-tuning (RL) or paraphrasing. The findings demonstrate the potential for malicious actors to circumvent classification. Further, the need for ongoing research into more robust and adaptive detection mechanisms is underlined.

\thispagestyle{fancy}
\renewcommand\headrulewidth{0pt}
\fancyhf{}

\section{Related Work}

Since this work combines different fields, this section is subdivided: Firstly, ways to automatically generate text using LLMs are explained (Sec.~\ref{sec:rel_textGeneration}). Secondly, relevant datasets respectively benchmarks are brought in as well (Sec.~\ref{sec:rel_dataSets}). Thereafter, detection methods for identifying LLM content are introduced (Sec.~\ref{sec:rel_detectTechniques}). Additionally, possible ways how to bypass those classifiers are discussed.

\subsection{Automatic Text Generation}\label{sec:rel_textGeneration}

Various approaches have been employed in developing language models capable of generating text. The most prevalent architecture is based on transformer models, which include GPT and its predecessors like GPT-Neo-125M, GPT-Neo-1.3B, GPT-Neo-2.7B \parencite{gpt-neo}, GPT-J-6B \parencite{gpt-j}, OPT-125M, OPT-350M, OPT-1.3B, OPT-2.7B \parencite{zhang2022opt}, and GPT-2 \parencite{radford2019language}. Other variants encompass Instruct-GPT \parencite{ouyang2022training}, Google's T5 models \parencite{2020t5} as well as their Gemma series \parencite{gemma_team_gemma_2024}, Metas's Llama series \parencite{dubeyLlamaHerdModels2024}, Mistral AI's Mistral \parencite{mistral7b} and Mixtral \parencite{mixtralexperts} series, the Qwen models \parencite{qwen2} from Alibaba Cloud, to name just a very few.

These LLMs are often referred to as \textit{stochastic parrots} \parencite{bender2021stochasticparrots} since they generate sentences by predicting the next word based on probability distributions. The selection of the token with the highest probability is known as \textit{greedy search}. As humans do not always choose the most probable sequence of words, this purely deterministic approach does not capture the variability in human language \parencite{TheCuriousCase}. To introduce randomness, pure random \textit{sampling} selects tokens proportionally to their probabilities, increasing diversity but eventually reducing coherence \parencite{bengio2003neural}. Intermediate approaches have been developed to balance determinism and diversity. These include \textit{typical sampling}, which prioritizes tokens near the mode of the probability distribution, and \textit{top-k sampling}, which restricts choices to the \textit{k} most probable tokens \parencite{meister2022locally, fan2018hierarchical}. Alternatively, \textit{nucleus sampling} dynamically adjusts the candidate set to include only tokens whose cumulative probability remains below a threshold \textit{p} \parencite{TheCuriousCase}. Each method reflects trade-offs between control, coherence, and creativity.

Closely associated with \textit{creativity} is also the \textit{temperature}. The parameter $\tau\in [0,2]$ modulates the sharpness of the probability distribution, where higher temperatures ($\tau>1$) flatten the distribution, allowing for more diverse and creative outputs but increasing the risk of incoherence. Conversely, lower temperatures ($\tau<1$) sharpen the distribution, focusing on high-probability tokens and yielding more deterministic but less creative text. Different sampling strategies might be combined with $\tau$, tailoring text generation to specific requirements.

\subsection{Datasets \& Benchmarks}\label{sec:rel_dataSets}

A direct comparison of human versus machine-generated content is provided by the \textit{Human ChatGPT Comparison Corpus} \parencite[HC3;][]{ChatGPTvsHuman}. This dataset consists of questions regarding various topics, each answered by an LLM versus real humans.

For question-answering research, Googles \textit{Natural Questions} corpus \parencite[NQ;][]{NaturalQuestionsGoogleBenchmark} serves as a benchmark. This dataset consists of real user search requests. Using CNN and Daily Mail articles, another question-answering corpus is provided by \textcite{cnnDailyMailDataset}.

The \textit{Corpus of Linguistic Acceptability} dataset \parencite[CoLA;][]{CoLA} comprises sentences that have been labeled as either grammatically acceptable or unacceptable by human annotators. It might be used for training LLMs such as DeBERTa-v3-large \parencite{he2021debertav3}, serving as a classifier for evaluating linguistic acceptability.

\subsection{Detection Techniques}\label{sec:rel_detectTechniques}
In the social media context discussed here, analyses are built on the users' \textit{behavior} or \textit{content}. While focusing on content specifically (i.e., posts on X, formerly Twitter), the reader is referred to the literature review by \textcite{alothali-2018} for deeper insights into the behavior-based approach. 

Various focus and content-independent detection methods exist. According to \textcite{orabi-2020}, these can be categorized as
\textit{graph-based} approaches \parencite[e.g.,][]{abou2020botchase}, \textit{crowdsourcing} techniques \parencite[%i.e., manually bot identification, 
e.g.,][]{wang2012social}, \textit{anomaly} detection methods \parencite[e.g.,][]{nomm2018unsupervised}, and \textit{machine learning-based} approaches \parencite{alothali-2018}. This categorization by \citeauthor{orabi-2020} can be broken down to \textit{machine-} versus \textit{human-based techniques} since detection is performed either manually (e.g., bot identification) or by machines, i.e., based on ML or pure statistical anomalies. 

\subsubsection*{Human-based Detection}

Human performance can be examined in experimental settings. Here, human versus machine-generated texts are presented to the participants (independent variable). The subjects, who do not know the origin of the texts, are asked to identify machine-generated ones. Based on a representative random sample, this procedure allows the estimation of general human accuracy, false alarm rate, etc. (dependent variables). These scores can not only be compared to those resulting from computer-based classifiers but can also be used to identify characteristics, i.e., what humans or machines tend to prioritize and what makes them more likely to fail.

Demonstrated that way, humans tend to prioritize the semantic coherence of the text. At the same time, machine-based detection methods emphasize statistical properties such as word probabilities and sampling schemes \parencite{DBLP:conf/acl/IppolitoDCE20}.

Humans seem to know how and what other humans would talk about compared to machines. The fact that machines can mimic these expressions and content was used by \textcite{jakesch_human_2023}: In their experiment, participants could not distinguish between AI-generated and human-written self-presentations misled by the AI's usage of first-person pronouns, contractions, or family topics.

It can be concluded that time and resource-intensive human-based detection does not lead to better results since humans can easily be deceived by using the above-mentioned factors. This has been confirmed by \textcite{DBLP:conf/emnlp/DuganIKC20}, who introduced a tool for assessing human detection capabilities. Their demonstration of how easily humans can be deceived underlines the importance of machine- and statistical-based detection.

\subsubsection*{Machine-based Detection}

Many text generation models leave behind specific artifacts whose occurrence is extremely unlikely compared to human text \parencite{DBLP:conf/acl/TayBZBMT20}. Those probabilities or word frequencies can be examined using statistical methods (see below). Since a manual calculation is theoretically possible here, these methods can be considered as \textit{simple}. These are to be distinguished from those methods that require the use of ML models, such as LLMs, which are therefore considered separately. 

Models using the transformer architecture can be used for language generation and detection. The MultiNLI benchmark \parencite{MultiNLI} allows a performance comparison regarding detection: Here, BERT-large achieves an accuracy of 88\% \parencite{BERT-Large}, while RoBERTa and DeBERTa score 90.8\% \parencite{RoBERTa} and 91.1\% \parencite{DeBERTa}, respectively.

However, for every generator release, BERT-based classifiers such as RoBERTa must be trained again. Therefore, zero-shot classifiers like DetectGPT become quite handy \parencite{mitchell_detectgpt_2023}. These classifiers only require two models: A duplicate of the model to be tested and a second language model introducing random permutations into the test string. However, the permutation models can not handle short inputs due to their working principle. 

Paraphrasing outlines the weak spot of the discussed classifiers. AI-based paraphrasing can be detected successfully, as shown by \textcite{li_spotting_2024}. However, this technique focuses solely on paraphrasing detection. Thus, the model has no ability to tell if the original text was human- or machine-generated. Therefore, paraphrasing can still serve as an effective bypassing technique. 
For instance, this technique is used by \textcite{krishna_paraphrasing_2024} and enhanced by \textcite{sadasivan_can_2024} using recursive paraphrasing. While \citeauthor{krishna_paraphrasing_2024} introduced the T5-based DIPPER as their own paraphrasing model, \citeauthor{sadasivan_can_2024} used DIPPER and other existing models combined without any model training or fine-tuning. \textcite{sadasivan_can_2024} claim their model generates paraphrases with high \textit{text quality} and \textit{content preservation} based on human ratings on a five-point Likert scale. 

The model's output can be altered not only through a separate paraphrasing model but also through a reinforcement learning approach that directly modifies the generative model. This method is inspired by the paper \textit{Fine-Tuning Language Models from Human Preferences} by \textcite{DBLP:journals/corr/abs-1909-08593-neu}. 

Another approach focuses on circumventing DetectGPT \parencite{mitchell_detectgpt_2023} using a paraphrasing model \parencite{krishna_paraphrasing_2024}. The countermeasure suggested by \citeauthor{krishna_paraphrasing_2024} detects AI-generated content by comparing it to a database of AI-generated texts. This solution has two significant drawbacks: First, the provider of this AI model must establish such a service, and second, custom fine-tuned private-run models are inaccessible \parencite{kucharavy_llm_2024}.

When the underlying generative model is known, it can be modified and hence serve as a classifier after further training \parencite{zellers_defending_2019}. \citeauthor{zellers_defending_2019}, who originally built up their model for fake news generation, proved that this model is also most effective in detecting their own fake news. This implies that detectors originating from the generative model itself are better at detecting artificially generated fake news than standard classifiers. Both fake news generators and fake news detectors are combined by \textcite{henrique_stochastic_2023} as generators and discriminators in the form of a generative adversarial network (GAN) to demonstrate the attack against a classification model. 

\subsubsection*{Statistical-based Detection}

After ChatGPT's introduction \parencite{gpt-intro-2022}, the frequency of certain words, such as \textit{intricate}, \textit{meticulously}, \textit{commendable}, or \textit{meticulous}, has changed significantly in academic literature \parencite{gray_chatgpt_2024}. In particular, these words increased by 50\%, and others, such as \textit{innovatively} even by 60\% compared to pre-2022 levels; for a complete list of words and statistics, see the dataset by \textcite{gray_dataset_2024}. While one of these so-called \textit{group-1} words may have been used by chance, the multiple use of \textit{group-1} words within a text is so unlikely that it can be considered GPT-generated. \textcite{gray_chatgpt_2024} estimates 1\% (60.000 articles) of all 2023 publications to be machine-generated and even predicts a further increase in 2024.

Several other rule-based models exist, specially designed to identify automatically generated texts. These rely on improbable word sequences and grammar \parencite{DBLP:journals/jasis/CabanacL21} and similarity measures such as word overlap \parencite{harada2021discrimination}.

A statistical-based countermeasure is the usage of watermarks in generated texts, as suggested by \textcite{kirchenbauer_watermark_2023}. This approach adds a constant $\delta$ to a green list of tokens while they are autoregressively sampled. So, some words appear slightly more often than others, although they are in the right position and do not jeopardize the sentence's grammar or meaning. 
Due to the potential of paraphrasing, detecting machine-generated text using watermarks has limitations. Removing one-quarter of the watermark tokens can be enough to evade the detection \parencite{kirchenbauer_watermark_2023}. Additionally, the generation itself infers with the sampling process of the language model by withholding certain tokens and strengthening others.
\section{Experiment 1: Evasion of Shallow Detectors}\label{sec:exp1}

Various variables may affect the accuracy (ACC) of a detector, above all, the type of detector itself. For example, transformer-based detectors will generally perform better than shallow ones. Although shallow detectors are less accurate, they are faster and use fewer resources, making their use in practice seem realistic (e.g., real-time detection in large social networks). Serving as a benchmark, the focus of the first experiment is thus on detectors such as Naive Bayes combined with Bag-of-Words (BoW). 

The ACC of shallow detectors is considered as the dependent variable. From the related work, it can be deduced that the ACC should be influenced by the now explained independent variables (i.e., LLM type, resp. size, sampling strategy, resp. size, as well as temperature). 

The OpenAI GPT family has demonstrated that increasing the number of model parameters leads to significant improvements in language generation. Larger models, however, are particularly effective in generating longer texts due to their extended context windows. 
Given that the current scope is generating short tweets, varying the model size will answer the question of whether larger models also beat smaller ones when the generated output is short.

Further, the influence of the sampling strategy on the LLM token selection process is examined. It is assumed that selecting the most probable token at each step (i.e. \textit{greedy search}) should be the easiest to detect due to determination. Conversely, pure random sampling should be the most difficult to detect for machines. However, the texts generated in this way are unusable due to their low coherence and, hence, noticeable to humans. Consequently, these two sampling strategies only set the borders for the remaining ones, which are of true relevance. These either limit the number of high-probability tokens $k$ (\textit{top k sampling}), consider a cumulative probability threshold $p$ (\textit{nucleus p sampling}), or are based on the conditional entropy of token sequences (\textit{typical sampling}). Although it is unclear which of these strategies will perform best, previous comparisons by \textcite{TheCuriousCase} imply that \textit{nucleus sampling} should be prior. However, it is unsure whether these results are generalizable to the present context, especially in combination with the other independent variables.

From the possibility of statistical detection explained above, it can be deduced that the sample size should also have an effect in addition to the type of sampling. It is hypothesized that as the sample size increases, the representativeness is enhanced, and thus, the detector's accuracy is also improved. Conversely, the ACC decreases for smaller sample sizes.

Last but not least is the influence of temperature, which adjusts the probability distribution for selecting the next word in the sequence to be predicted. Its definition implies that higher temperatures ($\tau>1$) lead to more unusual texts, which are more difficult for machines to detect but might be recognizable to humans due to incoherence. Conversely, a low temperature ($\tau<1$) leads to high coherence but also determinability and is, therefore, easily recognizable by machines. Since these assumptions result from the definition of the parameter itself, a moderating effect of temperature over the other factors is assumed. In other words, the effects mentioned should be additive. At least, no indications in the literature make the interaction of a specific model-sample-temperature combination seem plausible.

\subsection{Methodology}

\begin{figure}[t] %[ht]
    % \centering
    {\caption{Data Pipeline used for Modeling}
    \label{fig:flowchart}}
    \FIG{\includegraphics[width=1\textwidth]{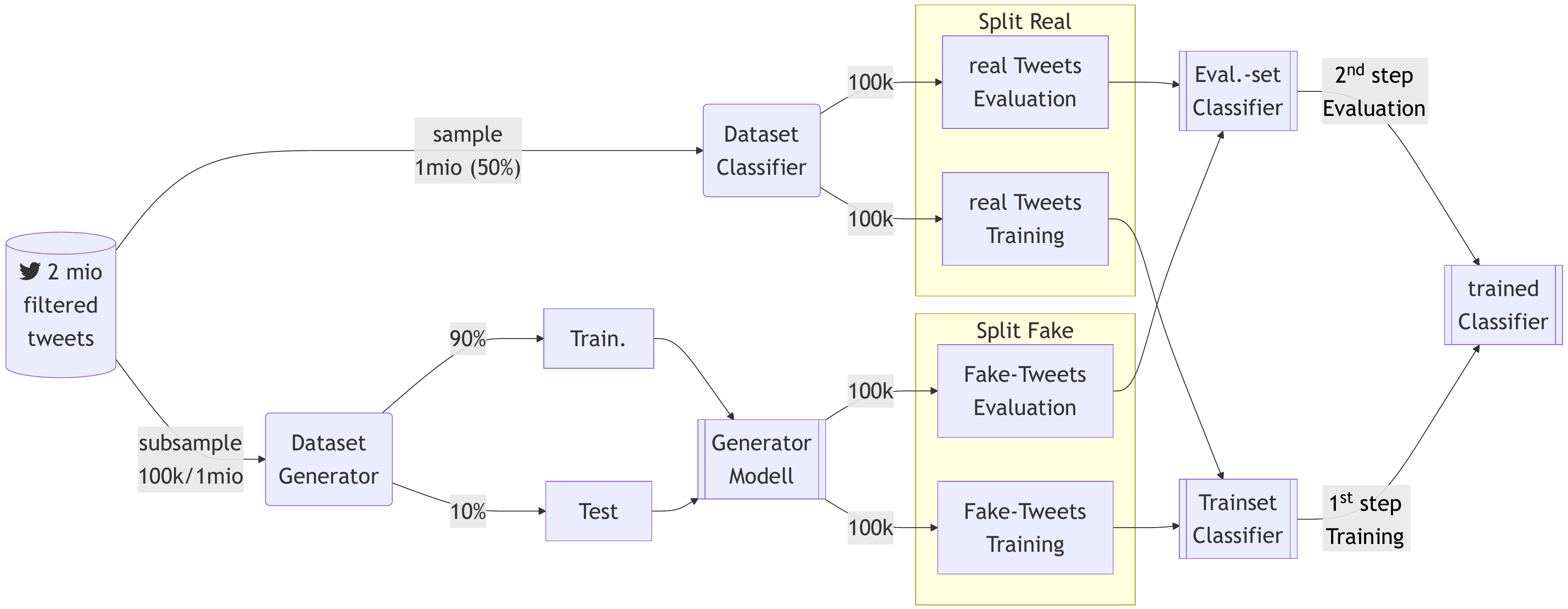}}
    \flushleft
    \footnotesize
    \emph{Note:} A comprehensive filtering policy was used (e.g., only tweets from verified users below the average amount of daily tweets; English language; no retweets or quoted tweets, etc.). Dataset yield as the basis for classifier and generator, respectively. Several models were tested (e.g., pre-trained GPT versions). 
\end{figure}

The procedure involved several steps: A dataset with real tweets was first filtered to ensure that only real tweets were involved (Sec.~\ref{sec:dataset_exp1}). This partial dataset was used to fine-tune various LLMs (Sec.~\ref{sec:generators_exp1}). The dataset was then completed using LLM-generated synthetic “fake” tweets (Sec.~\ref{sec:synthetic_tweets}), yielding from different hyperparameter combinations (Sec.~\ref{sec:parameter_grid}). A classifier was then trained with both real and fake tweets (Sec.~\ref{sec:detector_exp1}). Using a grid-based approach, the effect of the independent variables (i.e. hyperparameters) could be measured as a result of the classifier evaluation (Sec.~\ref{sec:results_exp1}). The multi-step procedure requires the data to be split several times to provide the different models with unseen data, as depicted in Figure~\ref{fig:flowchart}.

\subsubsection{Dataset}\label{sec:dataset_exp1}

The dataset employed in this study comprises tweets collected between January and February 2020. Due to the noisy nature of the raw data, a comprehensive set of filtering policies was implemented to refine the dataset. 
The following measures secure a clean dataset in one language.

In the first step, the dataset was restricted to tweets composed in English from authors with less than 100.000 followers. This measure excluded accounts from companies, sports teams, celebrities, etc., primarily used for advertising.
Moreover, non-truncated tweets were selected to ensure the text was fully available, which is essential for training.
To guarantee a wide variety of content, follow-up tweets, replies, quotes, and retweets were discarded. Further, only tweets from users who sent less than 20 tweets per day made it into the dataset. 

After applying the mentioned filtering steps, the dataset contained approximately 2 million tweets from 136.450 verified accounts. These are set to be real tweets exclusively. For adding fake tweets to the partial dataset, it was split first for training various LLMs. The dataset was then completed using LLM-generated synthetical fake tweets. In the next step, the dataset was split up again in order to train a classifier and check its ACC accordingly.

This procedure is illustrated in Figure~\ref{fig:flowchart}.

\subsubsection{Generative Models}\label{sec:generators_exp1}

Different open-source language models based on the GPT and OPT architecture are fine-tuned for later tweet generation. Particularly GPT-2 \parencite{radford2019language}, GPT-J-6B \parencite{gpt-j}, GPT-Neo-125M, GPT-Neo-1.3B, and GPT-Neo-2.7B \parencite{gpt-neo}. 
Furthermore, several OPT variants (125M, 350M, 1.3B, and 2.7B), as referenced in the work of \textcite{zhang2022opt}. 

\subsubsection{Synthetic Tweets}\label{sec:synthetic_tweets}

Using a pre-defined sampling strategy and temperature setting, each model generated two sets of 10.000 synthetic tweets for training and evaluation of the detection models, respectively. 

Examples are provided in Table~\ref{tab:methodology-example_tweets}.

\begin{table}[b]
\tabcolsep=2pt%
\TBL{\caption{Example Tweets generated with different Models\label{tab:methodology-example_tweets}}}
{\begin{fntable}
\begin{tabular*}{\textwidth}{@{\extracolsep{\fill}}ll@{}}\toprule%
\TCH{Model} & \TCH{Tweet} \\\midrule
        GPT-- & \\
        \hspace{1em} Neo-125M & {\fontfamily{cmss}\selectfont\small
Kobe says new coronavirus warning on plane is too difficult to understand.} \\
        \hspace{1em} Neo-1.3B & {\fontfamily{cmss}\selectfont\small The new album is out now; make sure you have the album download code for free.} \\
        \hspace{1em} Neo-2.7B & {\fontfamily{cmss}\selectfont\small \#ValentinesDay: Today is the day to celebrate the greatness of yourself. And to\dots}\smash{\footnotemark[1]} \\ % appreciate
        \hspace{1em} J-6B & {\fontfamily{cmss}\selectfont\small "This is how we play games!" Let’s hear "The Box" tonight with @OzzyOsbourne\dots}\smash{\footnotemark[1]} \\

        OPT-- & \\
        \hspace{1em} 125M & {\fontfamily{cmss}\selectfont\small I'm sure a few will be added in a future update as part of the "Duke" legacy.} \\
        \hspace{1em} 350M & {\fontfamily{cmss}\selectfont\small Good luck on the final stage of your tour!} \\
        \hspace{1em} 1.3B & {\fontfamily{cmss}\selectfont\small Rangers' Henrik Lundqvist: "I'm not even thinking about' the trade rumors, \dots"}\smash{\footnotemark[1]} \\ % says the goalie
        \hspace{1em} 2.7B & {\fontfamily{cmss}\selectfont\small A very cold, chilly \#day for \#Lincoln and \#Omaha \#MorningWeather}\\
\botrule
\end{tabular*}%
\footnotetext[]{\emph{Note:} Tweets generated with temperature $\tau=1.0$ and top-50 sampling. $^1$ Example trimmed due to excessive length.}
\end{fntable}}
\end{table}

\subsubsection{Parameter Grid}\label{sec:parameter_grid}

The independent variables are hyperparameter combinations of a parameter grid, including \textit{temperature}, \textit{sampling scheme}, and \textit{-size}. \textit{Temperature} is varied from $\tau_{min}=0.8$ to $\tau_{max}=1.4$ in steps of $0.2$ (the results show that this part of $\tau\in [0, 2]$ is sufficient, however, for the most analysis the borders were increased to $\tau'_{min}=0.6$, using steps of $0.1$). Five sampling schemes (greedy search, typical-, top k-, nucleus- and random sampling) with four sampling sizes (1k, 10k, 50k, and 100k) were used. The whole parameter grid is used for fine-tuning a  GPT-2 (1.5B) model and training the classifier, respectively.

A subsequent analysis is conducted to check if architecture and parameter size have an effect. The design is reduced to the most promising sampling scheme $\times$ size combination (i.e., easily detectable combinations are excluded from the design). Hence, the outstanding combination was tested for a branch of nine models with six different parameter sizes, all on the given temperature range. The used models differed in architectures (OPT vs. GPT) and parameter size, ranging from 125M to 6B. 

\subsubsection{Detector}\label{sec:detector_exp1}

A Naive Bayes classifier using BoW features was applied to detect synthetically generated tweets. This classifier has been chosen for its simplicity and short training time.

A grid-based approach was employed to train the classifier with various parameters 
(temperature, sampling strategy resp. size, model architecture resp. size) all of which were explained in detail above. 

\begin{figure}[t]
{\caption{%\small 
Human-based against machine-based Word Probability Distributions}
\label{fig:qqplotone}}
\FIG{\includegraphics[width=0.95\textwidth]{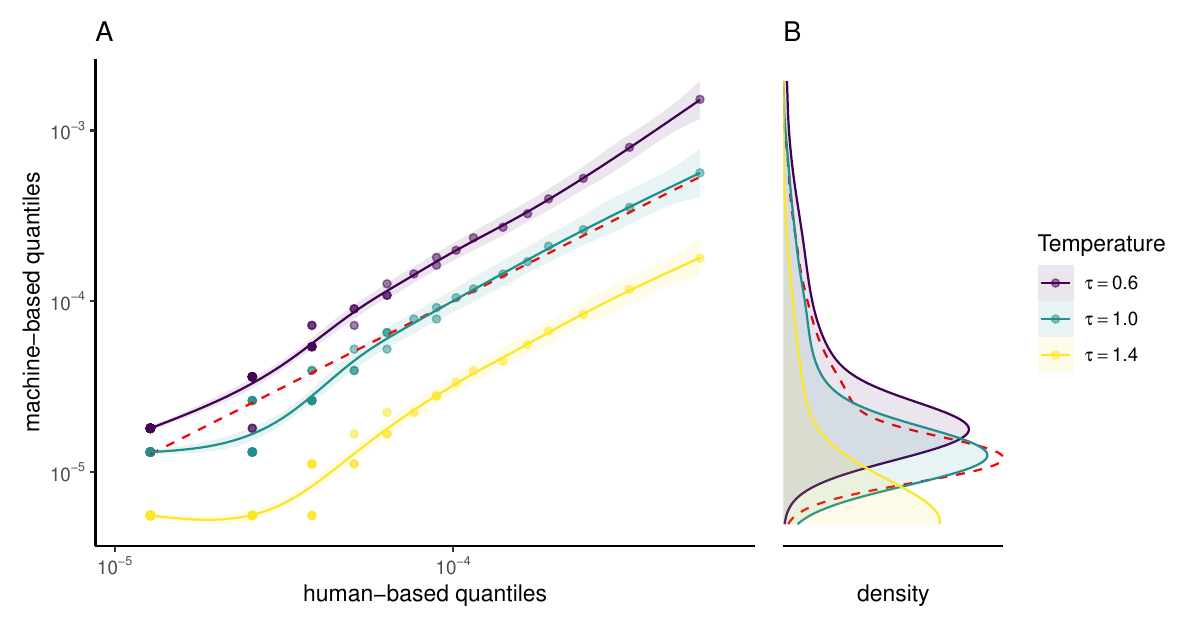}}
    \flushleft
    \footnotesize
    \emph{Note:} Logarithmic scaling applied to quantiles on both axes. Machine-based quantiles result from a GPT-2 (1.5B) model using random sampling with a sampling size of 10k. \textbf{\sffamily A}. Comparison of machine against human-based word probability distributions (red dashed line marks theoretical perfect mapping). \textbf{\sffamily B}. Density distributions reflect the effect of temperature (red dashed line marks empirical human density distribution).
\end{figure}

\subsection{Results}\label{sec:results_exp1}

\begin{figure}[t]
{\caption{Detectionrates by Temperature for Sampling Sizes, Methods, and Generator Models}
\label{fig:resultsExp1}}
\FIG{\includegraphics[width=.99\textwidth]{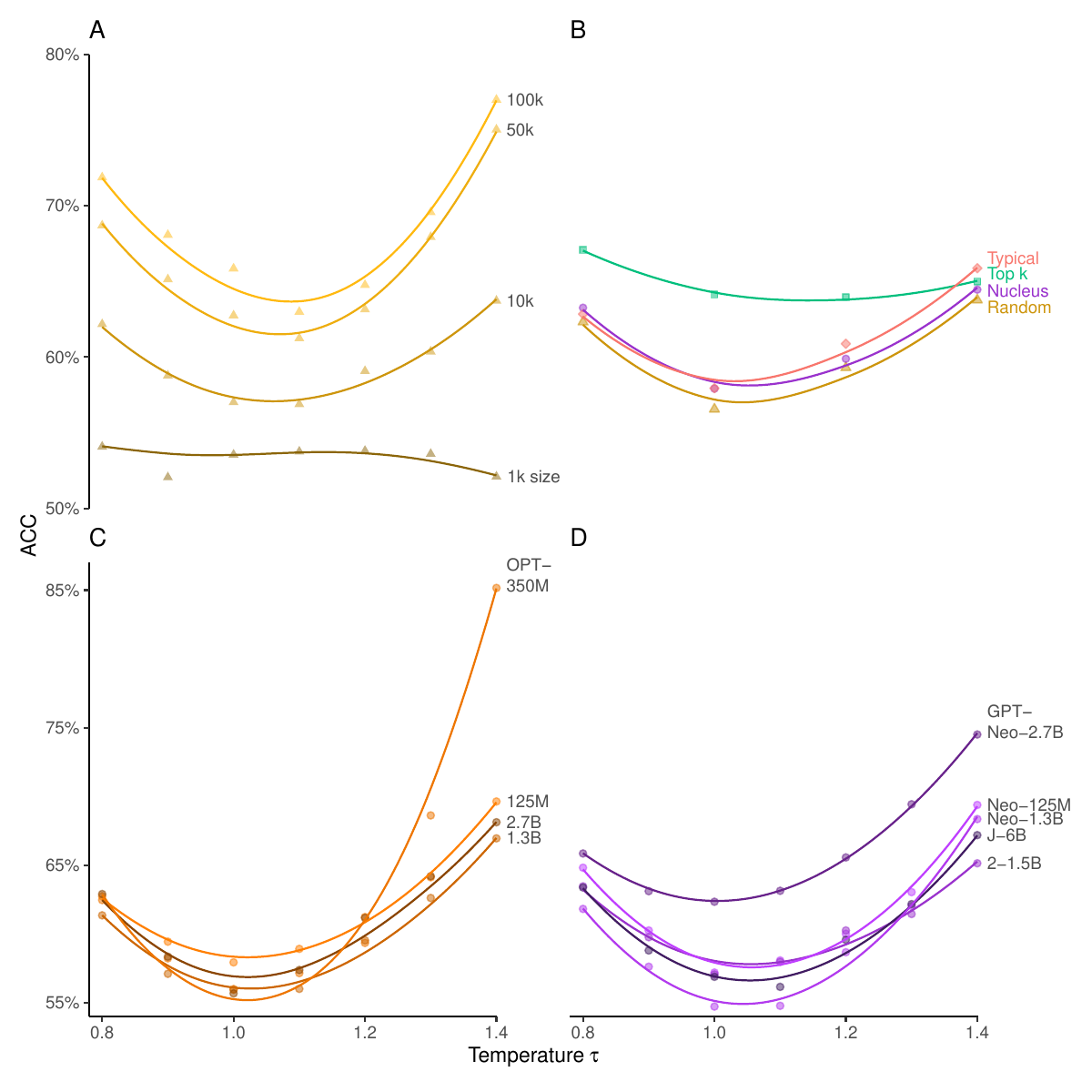}} % images/hx_plt_4.pdf
    \flushleft
    \footnotesize
    \emph{Note:} Comparison of ACC for varying temperatures $\tau$ and \textbf{\sffamily A}. different sampling sizes (1k--100k tweets) using \textit{random sampling} and \textbf{\sffamily B}. different sampling types, i.e.,\textit{ typical--}, \textit{top} $k = 100$, \textit{nucleus} $p=0.95$ and pure \textit{random} sampling, all with 10k sampling size. \textit{greedy search} not depicted here, since it always leads to $ACC>99\%$. Both sampling sizes and strategies result from the same GPT-2-1.5B model, whereas panels \textbf{\sffamily C}. depicts results for OPT-- and \textbf{\sffamily D}. for GPT model architectures with various parameter sizes. Sampling sizes and model parameters are reflected by color shading (i.e., the darker, the bigger). For all panels, small ACC values indicate a better performance of the generating model. 
\end{figure}

By definition, the temperature modification for LLM tweet generation led to modified word probability distributions accordingly. A comparison of human against machine-based quantiles revealed notable differences for temperatures of $\tau\ne 1$. This characteristic is illustrated in Figure~\hyperref[fig:qqplotone]{2A}, where the frequency distribution is less visible due to logarithmic scaling. Visualized by plotting the density distribution accordingly, deviations are clearly notable for density distributions resulting from $\tau\ne 1$ (v.v. less for $\tau= 1$, cf. Fig.~\hyperref[fig:qqplotone]{2B}). Those depicted differences make it easy to detect variations for statistical-based classifiers, such as Naive Bayes. 

Across models, sampling sizes, and temperature values, the ACC for different sampling strategies was maximal when using greedy search, as expected, and minimal when using random sampling. In the range of both strategies, nucleus sampling yielded the best results, followed by typical- and top k sampling with similar results.  
For all sampling schemes, a U-shaped result pattern emerged regarding increasing temperature values, as depicted in Figure~\hyperref[fig:resultsExp1]{3B}. The same pattern is visible for different sampling sizes, where larger samples did not only lead to higher detection rates but also to higher gradients for temperatures diverging from 1 (cf. Fig.~\hyperref[fig:resultsExp1]{3A}). 

Across temperatures, the different models did not show a clear trend, neither in the model architecture nor in size. Particularly, both architectures lead to similar results, $M_{ACC}\approx 61.80\%$ for both types. Using Spearman's $\rho$ showed rather a small positive, then a negative correlation between ACC and parameter size, $\hat{\rho} = .14$ for GPT-- and $\hat{\rho} = .13$ for OPT models (both not significant).
This is visualized for different OPT-- (Fig.~\hyperref[fig:resultsExp1]{3C}) and GPT models (Fig.~\hyperref[fig:resultsExp1]{3D}), using nucleus sampling with $p= 0.95$ and a sample size of 10k. A repetition of the experiment revealed that outliers (as shown here for OPT-2.7) were unsystematic.
Notably, the same U-shaped pattern as described for sampling sizes and strategies is also visible across models: Centered at $\tau\approx 1$ where ACC is minimal, diverging temperatures lead to an increase in ACC.

\subsection{Discussion}

Experiment 1 focused on attacking shallow learning detectors to obtain an ACC baseline measure. Therefore, synthetic tweets were generated with the manipulation of several independent variables. The ACC of these detectors was then tested using a Naive Bayes classifier combined with BoW.

When the temperature was set to $\tau=1$, the detection rate was minimal. Notably, there was no (curvi-)linear relationship between temperature and ACC but a U-shaped one centered around $\tau\approx1$. At this point, human and machine-generated word probability distributions were most similar. Vice versa, $\tau\ne1$ led to easily distinguishable distributions and, hence, to higher detection rates. This effect increased the more $\tau$ diverged from its centers. Also, the effect became stronger when sampling sizes increased (v.v. less to not visible for small samples, e.g., 1k). Regarding sampling methods, the easiest to detect was greedy search since it works by always selecting the most probable next token, introducing very little entropy (i.e., max. ACC>99\%). Comparably, the most difficult to detect was randomly sampling all possible next tokens, introducing more entropy and resulting in detection rates below 60\%. However, for the purpose of coherence, using different sampling schemes is advisable. Here, nucleus sampling led to the best results (i.e., the lowest ACC). Model architecture and size, however, play a minor role in detecting short texts (i.e., similar results across architectures and no clear trend of parameter size across models). While larger models beat smaller ones when the generated output is long, this effect is not true regarding short output text. One possible explanation of this is the fact that larger models can not display their strength regarding extended context windows when texts are already small in the first place. 

Taken all together, the results show, that shallow learning-based classifiers appeared to perform insufficient if the generative models produced texts with high entropy and similar word distribution to human texts. This can be traced back to their working principle since the BoW approach does not consider the position in which a certain word is located nor synonyms with the same meaning.

Shallow detectors are resource-sparing classifiers and, therefore, represent a realistic application example (e.g., monitoring big data streams). At the same time, these detectors can be evaded easily, as demonstrated here. More advanced, resource-intensive classification models are therefore used in the field. Those models, such as those based on BERT ones, push the boundaries of the here presented strategies. Hence, alternative evasion techniques are needed for transformer-based detectors, as presented in Experiment 2.
\section{Experiment 2: Evasion of Transformer-based Detectors 
}\label{sec:transformerBasedDetectors}\label{sec:exp2}

Given their superior performance in text classification, transformer-based models are increasingly preferred for deployment in production environments. Hence, the previously applied evasion tactics become insufficient. This raises the question of whether more advanced classification models can also be evaded.

To examine this question, the former described procedure was adopted accordingly on both sides by replacing the shallow classifier with a transformer-based one and adjusting the evasion strategy as well.

Unlike shallow learning and other deep learning algorithms, transformers make one major improvement: Their self-attention capability, formed by self-attention layers. These layers facilitate the mapping of tokens into a vector space in relation to their surrounding tokens, resulting in more contextualized representations.

Simple parameter tuning is insufficient to bypass these attention-equipped transformer models. Hence, it is vital to use more sophisticated methods. One of those might be reinforcement learning (RL). By using RL, a model can be guided towards creating a desired output. For example, it is possible to change the sentiment of a text from \textit{negative} to \textit{positive}. Hence, the desired output can also be a non-detectable text. RL is, therefore, used here to bypass transformers.

However, reinforcement learning's unpredictable characteristic in finding strategies for reward maximization comes with the risk of the model learning introducing artifacts to bypass the classifier. To mitigate this risk, additional guardrails are applied. 

\subsection{Methodology}

The previously used procedure was adapted to evade transformer-based detectors. Firstly, the BoW text encodings were replaced by a transformer-based classifier (Sec.~\ref{sec:detector_exp2}). Secondly, to bypass this classifier, reinforcement learning was used (Sec.~\ref{sec:ReinforcementLearning}). To further stabilize the learning process, the RLs reward function was divided into a) the classical evasion reward (Sec.~\ref{sec:Rewardfunction}) and b) further constraints. The later are not only vital for stabilization, but also hinder the RL process to find undesirable evasion tactics (e.g., introducing artifacts such as extensive usage of special characters).  

The procedure was applied to a second scenario, i.e., the generation of fake news, to demonstrate that this methodology can be adapted to other domains. In this regard, a new dataset was used (Sec.~\ref{sec:dataset_exp2}). Training and testing procedures % for this scenario 
were similar and are therefore not again described. %to those used in the previously elaborated fake tweet experiment, 
However, the linguistic refinement filters could be simplified to one single rule. %(Sec.~).
In particular, this rule ensured that generated texts had different starting phrases, thereby maintaining some level of diversity in the output.

\subsubsection{Dataset}\label{sec:dataset_exp2}

% \subsubsection{Environment and Dataset}
The used human dataset was the same as described in the previous Experiment (Sec.~\ref{sec:dataset_exp1}). To complement the dataset with synthetic tweets, a top-50 sampling with a temperature setting of $\tau=1.0$ was used. Additionally, the number of training samples for a BERT classifier was fixed to 100k compared to the BoW classifiers.

In order to assess the generalizability of the reinforcement learning approach, a second iteration of text generation using the CNN/Daily Mail dataset from \textcite{cnnDailyMailDataset} was conducted. 

\subsubsection{Detector}\label{sec:detector_exp2}

BERT was used as a primary reference model within the transformer family. This is due to its architecture, which reassembles the basics of its predecessors, such as RoBERTa and DeBERTa.

Given the considerable resources required to train an entire BERT model, fitting multiple models for the sake of hyperparameter optimization was discarded.
 
\subsubsection{Reinforcement Learning}\label{sec:ReinforcementLearning} 

The used hyperparameters, such as learning rate, mini-batch size, choice of the optimizer, and threshold for detecting linguistic acceptability, were based on literature recommendations. For the GPT-Neo-2.7B model, the linguistic acceptability threshold was reduced from 0.4 to 0.3, as larger models are more prone to manual interventions. Furthermore, the Adam optimizer was substituted with the Lion optimizer \parencite{Chen2024optimizationAlgo}, %\parencite{https://doi.org/10.48550/arxiv.2302.06675}, 
which has reportedly outperformed the former in certain scenarios.
During both the reinforcement learning and evaluation phases, the same sampling method was used to ensure consistency in results.  

The general RL procedure followed the three distinct stages as described by \textcite{vonwerra2022trl}: \textit{rollout}, \textit{evaluation} and \textit{optimization}. Since the predefined \textit{optimization} algorithm was not changed, only \textit{rollout} and \textit{evaluation} stages are described below in more detail. 

\paragraph{\textbf{Rollout}}
The first stage, named \textit{rollout}, entails the generation of synthetic tweets utilizing the language models described in Section~\ref{sec:generators_exp1}. %~\ref{sec:transferLearning}.
In this stage, the model is provided with the beginning tokens of an original tweet and is tasked with completing the sentence. Sometimes, the model generates entire tweets independently to mitigate the risk of overfitting short text fragments.

\paragraph{\textbf{Evaluation}}
During \textit{evaluation}, the generated texts are submitted to the BERT-based classifier described in Section~\ref{sec:transformerBasedDetectors}. 
%The second stage, \textit{evaluation}, uses the generated texts and submits them to the BERT-based classifier described in Section~\ref{sec:transformerBasedDetectors}. 
If the classifier recognizes the text as human-generated, the reinforcement learning algorithm receives a positive reward; otherwise, it receives a negative one.
In this context, raw logits have been found to yield optimal performance.

\paragraph{\textbf{Optimization}}
The final \textit{optimization} stage 
%The third and final stage revolves around \textit{optimization} and 
entails the computation of the log probabilities of the tokens to compare the current language model with a reference model. This step represents a critical element within the reinforcement learning framework proposed by \textcite{DBLP:journals/corr/abs-1909-08593-neu}, ensuring that the modified model does not overfit its generation process.

\subsubsection{Reward function}\label{sec:Rewardfunction}

\begin{figure}[t]
    {\caption{Reinforcement Learning Reward Calculation Procedure}
    \label{fig:rewardcalculation}}
    \FIG{\includegraphics[width=0.7\textwidth]{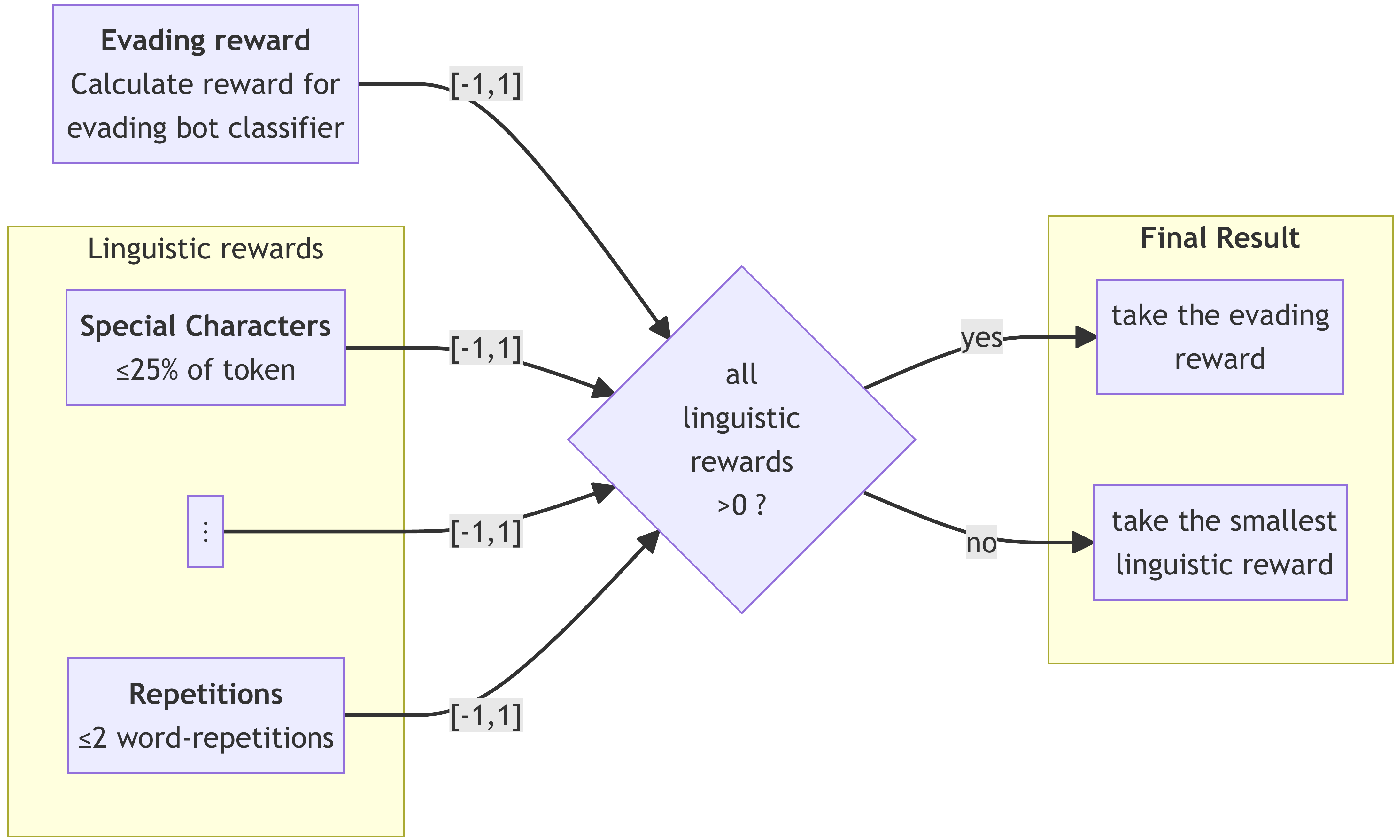}}
\end{figure}

In addition to the detector-based rewards, a carefully handcrafted reward function is introduced to further guide the text generation process.
This function penalizes generated texts that, while classified as human-like, fail to meet specific linguistic criteria. The reward calculation process is illustrated in Figure~\ref{fig:rewardcalculation} and consists of various rulesets, all of which will be outlined in the following paragraphs. If one or more of these linguistic rules are violated, the most severe penalty of all individual rules is applied. Conversely, should the model generate a synthetic text that satisfies all rules, the reward is equal to its evasion, as given by the detector model.

The optimization rules were developed by analyzing preliminary training runs, during which both request and response logs from the reinforcement learning process were examined. 
While reinforcement learning can operate without these rules, the resulting outputs are significantly less coherent. For instance, an unguided model may generate an output such as "Something for Administrator930 Macy’s Displays! RIP Family Members". While the detector algorithm did not classify this text as machine-generated, its core message is clearly questionable. Using the ruleset below, the restriction imposed by the linguistic acceptability rule would have prevented a positive reward from being assigned to this text.

The thresholds of the further introduced penalization rules are determined by observing the reinforcement learning logs. Looking at why a training run failed helps to iteratively build rule by rule instead of knowing all the constraints right from the beginning. Another important factor for choosing the right thresholds is the type of social media text. For example, a tweet might have more special characters and emojis than a book or newspaper text; consequently, a higher threshold is chosen for these measurements. 

For larger models, the reward associated with bypassing the detector algorithm can be multiplied by a scalar to prioritize the model's circumvention over producing grammatically or semantically refined sentences if necessary.

\begin{figure}[t]
{\caption{Ground-truth Distributions}
\label{fig:contraintsExp2}}
\FIG{\includegraphics[width=.99\textwidth]{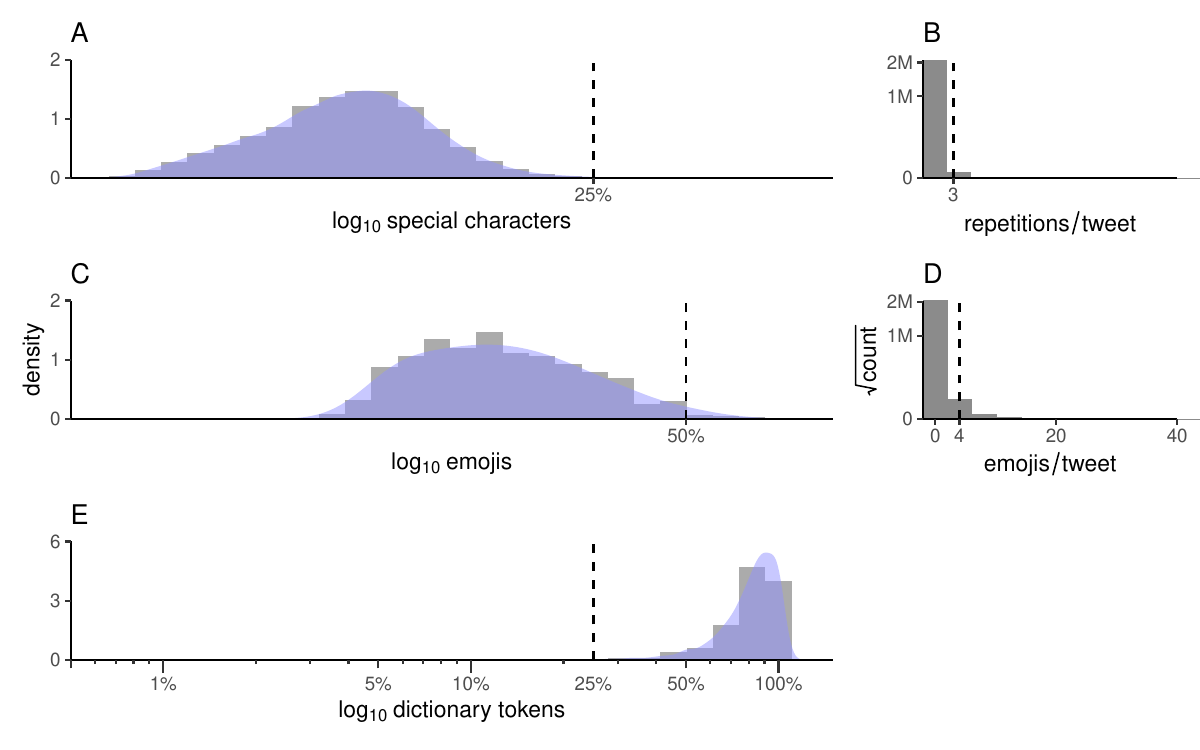}} % images/hx_plt_4.pdf
    \flushleft
    \footnotesize
    \emph{Note:} Constraints based on ground truth visualized for \textbf{\sffamily A}. the maximal proportion of special characters, \textbf{\sffamily B}. the number of repetitions per tweet, \textbf{\sffamily C}. the proportion of emojis and \textbf{\sffamily D}. the number of emojis per tweet as well as \textbf{\sffamily E}. the minimal proportion of tokens in a standard dictionary. For all plots, the dashed line depicts the cutoff values. For all proportions (1st col.), the x-axis is log-scaled for visualization purposes, for all frequencies (2nd col.), the square root of the actual numbers is depicted on the y-axis. 
\end{figure}

\paragraph{\textbf{Additional constraints}}
\subparagraph{\textbf{Special characters}} Texts containing more than 25\% special characters are penalized through a linearly decreasing negative reward, reaching the maximum penalty of -1 if the text consists entirely of special characters. Special characters include everything except Latin letters, numbers, and white spaces, while emojis are kept out of the calculation since they are treated in an extra rule.
The number of 25\% is backed up by the special character to all character ratio in the trainset of the generative model, as illustrated in Figure~\hyperref[fig:contraintsExp2]{5A}. This ratio might vary with changing text types, such as newspapers or books.

\subparagraph{\textbf{Repetitions}} Besides hallucination, the repetition problem is a well-known issue in natural language generation and is therefore already scientifically analyzed \parencite{repetitionProblem}. The idea behind the repetition penalty is to prevent the model from adopting this undesired behavior during the reinforcement learning phase. A text containing three or more instances of the same word is assigned a negative reward of up to -1 when the token is repeated eight times or more. In order not to prevent a natural text from being generated by this rule, the gold standard train corpus serves as a comparison where repetitions of over two times are very uncommon, as illustrated in Figure~\hyperref[fig:contraintsExp2]{5B}.

\subparagraph{\textbf{Linguistic acceptability}} Linguistic acceptability is evaluated using a DeBERTa-v3-large classifier \parencite{he2021debertav3}, which has been trained on the Corpus of Linguistic Acceptability (CoLA) dataset \parencite{CoLA}. CoLA comprises sentences that have been labeled as either grammatically acceptable or unacceptable by human annotators. The trained model assesses the grammatical acceptability of a given sentence, and if its score falls below the 40\% threshold, a negative reward is assigned. This reward again is linearly scaled and reaches a value of -1 if the acceptability score drops down to 0\%. 
For models with more than two billion parameters, the threshold is relaxed up to 30\% due to the increased training difficulty associated with larger models. The thresholds used in this evaluation were determined empirically rather than sourced from existing literature. Higher thresholds are typically recommended to enhance linguistic acceptability. However, excessively strict thresholds can impede the reinforcement learning process, as texts may be persistently classified as non-human, preventing the model from receiving positive rewards and hindering learning.

\subparagraph{\textbf{Dictionary}} Besides introducing artifacts such as special characters, the RL process could also lead to words that are not part of any dictionary. Since this is not uncommon for tweets, it is important to give the model a certain amount of freedom to introduce unknown words. However, this opportunity should not be used excessively. Therefore, the ratio between total words and unknown words of the original corpus is taken for comparison as illustrated in Figure~\hyperref[fig:contraintsExp2]{5E}. Because it rarely happens that less than 25\% of the words are part of a dictionary, a generation with a lower score generates a negative reward of up to -1 if none of the words are found in the dictionary.  

\subparagraph{\textbf{Word Emoji relationship}} Emojis are a common way to express emotions in social media. That's why they appear so often in social networks like X compared to newspapers or books. However, an overly excessive use of these expressions could also lead to a text losing its message and being undesirable to read. Therefore, the ground truth train dataset is once again consulted to find the maximum amount of desirable emojis within one tweet, as demonstrated in Figure~\hyperref[fig:contraintsExp2]{5C}. The threshold of giving a negative reward is reached once a tweet contains more emojis than words (50\%).

\subparagraph{\textbf{Number of Emojis}} 
The aforementioned word-emoji relationship might not be sufficient for longer texts since, in this case, many emojis are possible. A sentence of ten words could include ten emojis, which is a bit much. This is why, additionally to the sentence length, four or more emojis are given a negative reward. Such a penalty also aligns with the natural distribution of emojis among tweets (Fig.~\hyperref[fig:contraintsExp2]{5D}).

\subparagraph{\textbf{Repetition of the Query}} 
Although not as common as repetitions within a generated text, the repetition of the query is also a phenomenon that has been observed during RL training analysis. To conquer this flaw, repeating more than half of the query yields a negative reward. 

\subparagraph{\textbf{Special Tokens}} 
The use of special tokens, such as the beginning-of-sentence (BOS) and end-of-sentence (EOS) markers used within transformer models, is limited to two per tweet. The presence of each additional special token results in a negative reward of -0.4, with a maximum penalty of -1.

\subparagraph{\textbf{Same start}} Output diversity is essential when the model generates tweets without an input query. A negative reward is imposed if more than 10\% of the tweets in a training batch begin with the same word. This penalty increases linearly, reaching -1 if 20\% of the tweets start similarly.

\subparagraph{\textbf{Numbers at the start}} To prevent the model from learning to exploit number-based patterns to bypass the classifier, a penalty is given if generated tweets frequently start with numbers. 
If more than 10\% of tweets within a training batch exceed this limitation, the penalty is applied and scaled to a maximum once the frequency exceeds 20\%.

\subparagraph{\textbf{Unknown characters}} In some occasions, language models generate filler or unknown characters, typically caused by the occurrence of unknown characters included in the fine-tuning dataset. 
A starting penalty of -0.5 is given upon the first occurrence and decreases further for each consecutive appearance to prevent this undesirable behavior.

\subsubsection{Training Log}

To better illustrate the internal process, Table~\ref{tab:ReinforcementLEarning} provides a sample of the ongoing process, including logs, documenting queries, responses, and rewards. As outlined in Section~\ref{sec:ReinforcementLearning}, a positive reward is only assigned when both the query and response were classified as human-generated and none of the supporting rules produced negative feedback. In these cases, the reward relies solely on the BERT classifier's ability to detect generated content.

\begin{table}%[h]
\tabcolsep=0pt%
\TBL{\caption{Example of the Reinforcement Learning Process\label{tab:ReinforcementLEarning}}}
{\begin{fntable}
\begin{tabular*}{\textwidth}{@{\extracolsep{\fill}}llr@{}}\toprule%
\TCH{Query} & \TCH{Response} & \TCH{Reward} \\\midrule
        FINAL & {\fontfamily{cmss}\selectfont\small UPDATE: Chancellor says he has not shown anxiety} & $0.8019$\, \\
        {\fontfamily{cmss}\selectfont\small\textless |startoftext|\textgreater} & {\fontfamily{cmss}\selectfont\small see you this weekend! we'll be here again cider is back in v} & $0.9674$\, \\
        {\fontfamily{cmss}\selectfont\small\textless |startoftext|\textgreater} & {\fontfamily{cmss}\selectfont\small flush those lemons. This food ain’t good} & $0.9628$\, \\
        Ozzy Osbourne cancel & {\fontfamily{cmss}\selectfont\small s gig at studio theatre tonight over coronavirus 2020 forced} & $-0.0089$\, \\
        Today we'll know & {\fontfamily{cmss}\selectfont\small soon morning if there's blood somewhere in} & $-0.1756$\, \\
        The best \#adventure this & {\fontfamily{cmss}\selectfont\small year on 2020, dudes. Can we really} & $-0.2899$\, \\
        {\fontfamily{cmss}\selectfont\small\textless |startoftext|\textgreater} & {\fontfamily{cmss}\selectfont\small Yeah it’s weird how local commentators} & $0.9696$\, \\
\botrule
\end{tabular*}%
\footnotetext[]{\emph{Note:} The RL process steps are illustrated line by line.}
\end{fntable}}
\end{table}

\subsection{Results}

Across all evaluated BERT models, the mean $F_1$-score was quite high before the RL application, $M=0.94$ ($SD=0.01$). For the RL application, empirical analysis proved a learning rate of $5\cdot10^{-5}$ and a mini-batch size of 4, yielding the most optimal results. After this RL application, the mean detection rate decreased significantly to only $M=0.09$ ($SD=0.07$).

\begin{figure}[t]
    {\caption{$F_1$--Score Comparison}
    \label{fig:results_exp2}}
    \FIG{\includegraphics[width=0.95\textwidth]{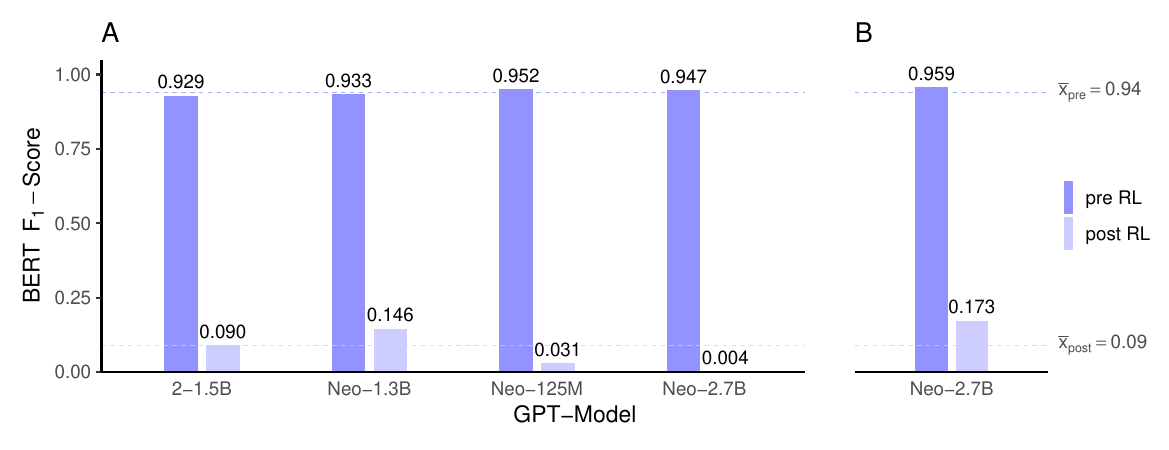}}
    \flushleft
    \footnotesize
    \emph{Note:} Results are depicted for \textbf{\sffamily A}. the used \textit{Twitter} dataset and also for \textbf{\sffamily B}. the \textit{CNN and Daily Mail} dataset as a representation of fake news. Dashed lines depict the mean $F_1$--Scores across models and datasets before and after RL. Line distance illustrates the huge RL effect of $d=-16.48$, $95\%\text{CI}[-24.14, -8.49]$, demonstrating transferability to different text domains.
\end{figure}

Further investigations regarding general applicability used the CNN/Daily Mail dataset. Despite the different domains, similar results were observed. Particularly, the detection rate decreased from $F_1=96\%$ to only $F_1=17\%$ after the RL application. Across applications, RL had a significant effect of $d=-16.48$, $95\%\text{CI}[-24.14, -8.49]$. According to \textcite{Sawilowsky2009rulesofTumb}, this effect can be interpreted as \textit{huge}. All results are summarized in Figure~\ref{fig:results_exp2}. 

\subsection{Discussion}

Experiment 2 focused on transformer-based detection mechanisms. The 2.7 billion parameter model, with its reduced linguistic acceptability threshold, yields notably superior results, although at the cost of reduced linguistic quality. Particularly, for LLMs such as GPT2 and GPT-Neo models of different sizes (125M--2.7B), all $F_1$-Scores exceeded 92\%. This high detection rate can be explained by the enhanced capabilities in spotting textual patterns.

Although transformer-based classifiers have demonstrated high reliability in distinguishing between real and machine-generated content, a fine-tuned reinforcement learning approach can effectively bypass these robust models. The application of RL had a \textit{huge} effect, decreasing the mean detection rates by more than 16 standard deviations to only $F_1=0.09$. This proves RL to be a reliable method for bypassing detection mechanisms. Furthermore, experiments adapted to the CNN/Daily Mail dataset demonstrate the applicability of this reinforcement learning approach to other text domains. Experiments with the four open-source models confirm that BERT classifiers can be bypassed using a reinforcement learning-based training methodology.

Despite the high detection rates, the potential for the BERT classifier to overfit to a specific generation method remains, which could lead to suboptimal performance when applied to other generative models.
\section{Experiment 3: Evasion of Zero-Shot-based Detectors }\label{sec:exp3}

Although the approaches of the first two experiments were successful, these attempts have several limitations. 

Primarily, the focus was solely on hiding the synthetic origin. Consequently, during RL, the model could change the content freely if it still made sense and was detected as human-written.

Additionally, the computational cost increases complementary with the parameter decrease of the model to be adjusted. This makes it increasingly difficult to run the RL approach from a hardware and stability perspective. 
Also, access to the original models is necessary to change them. Therefore, RL is limited to all open source models (i.e., not possible for, e.g., GPT 3 because the weights cannot be adjusted).

In order to compensate for these disadvantages and to create a general approach that is also valid for black-box models, an alternative procedure is proposed. This involves outsourcing the change to a new translation model that preserves the meaning while masking the origin.

One of the many advantages of transformers was the improvement of machine translation.

The most desired goal of translations is preserving the content while remaining linguistically well-written.
If transformers can be used to map from one language to another without changing the content (“translation”), it could also be possible to map from \textit{recognizable} to \textit{unrecognizable} with unchanged content.

For a transition to an experimental setting, maximum content similarity to the original and maximum unrecognizability are relevant. 
As with RL, unrecognizability is potentially in discrepancy with other influencing variables. While there is a risk of artifacts with RL, it can be assumed that content and sentence quality suffer here. Keeping the unrecognizability constantly high could lead to models, i.e., “translation” with inappropriate synonyms that are not detectable by classifiers but sound strange to humans. However, if all three influencing variables specified above, including sentence quality, are taken into account, this should lead to sentences being translated (paraphrased) in such a way that their LLM origin is no longer recognizable.

\subsection{Methodology}

To compensate for the previous language limitations, a new dataset consisting of LLM-answered questions was created (Sec.~\ref{sec:dataset_exp3}). The answers were paraphrased, and the results were filtered to obtain the highest coherence and similarity to the original answer while being less likely to originate from an LLM (Sec.~\ref{sec:filtering_exp3}). The thereby trained paraphrasing model (Sec.~\ref{sec:model_exp3}) was evaluated by using a different dataset and compared to reference models (Sec.~\ref{sec:evaluation_exp3}).

\subsubsection{Dataset}\label{sec:dataset_exp3}

\begin{figure}[t]
    \caption{Pipeline used for Trainingset Generation}
    \centering
    \includegraphics[width=0.9\linewidth]{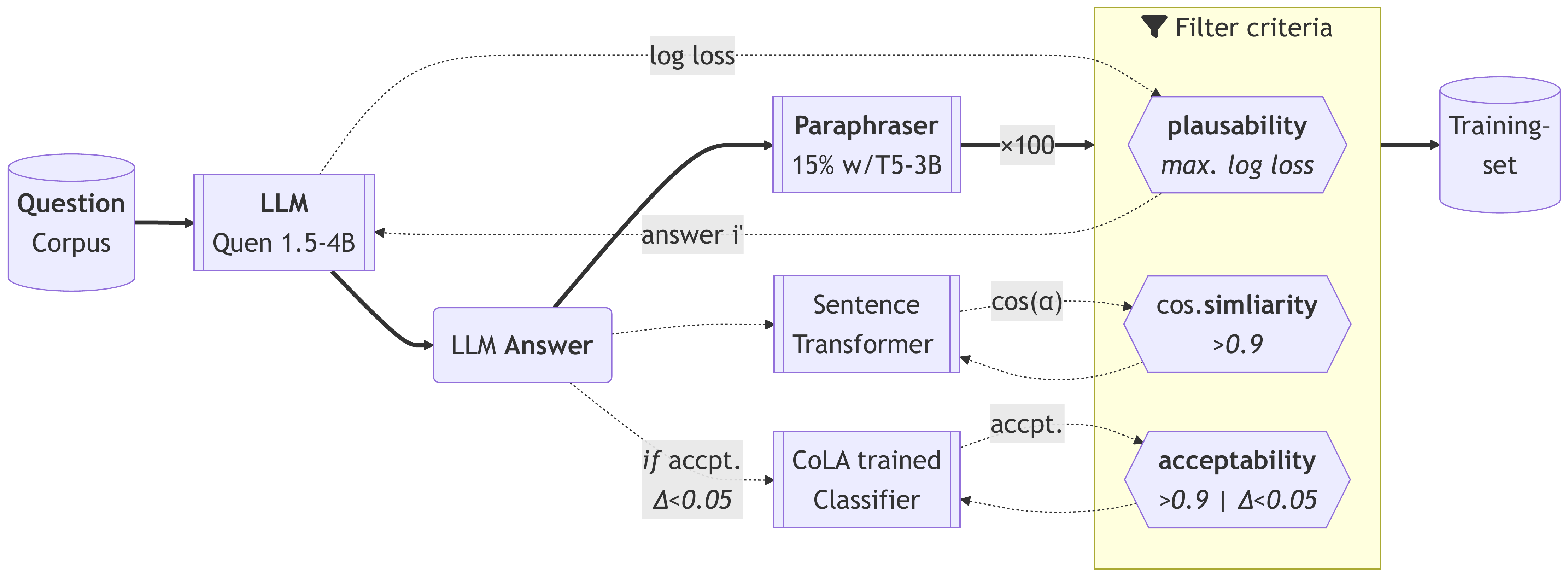}
    \label{fig:dataset-pipeline}
    \flushleft
    \footnotesize
    \emph{Note:} Question from Human ChatGPT Comparison Corpus \parencite{ChatGPTvsHuman} answered by Qwen1.5-4B-Chat \parencite{qwen_2023} and permutated by t5-3b \parencite{2020t5}. The Log loss (plausibility) is checked by the generative Qwen-model itself, while the similarity is checked using the sentence transformer all-MiniLM-L6-v2 \parencite{MiniLM}. The acceptability was checked by a DeBERTa \parencite{DeBERTa} model trained on the CoLA dataset \parencite{CoLA}.   
\end{figure}

This experiment utilized the \textit{Human ChatGPT Comparison Corpus} \parencite[HC3;][]{ChatGPTvsHuman} as the primary dataset. HC3 contains over 24,000 entries, each consisting of a question answered by both a human and ChatGPT. Instead of relying on the pre-existing GPT responses, new ones were generated using the \textit{Qwen 1.5-4B} model \parencite{qwen_2023}. This was necessary for the subsequent step, where the same model was used to calculate log loss for permuted answers. 
The permutation candidates for training a paraphrasing model were generated using the \textit{T5-3B} model without fine-tuning. %, which was trained on the Google PAWS dataset to produce paraphrases of English sentences \parencite[]{PAWSParaphraseAdversaries2019}.
Optimal permutation candidates for each answer were selected based on three criteria: \textit{Similarity}, \textit{coherence}, and \textit{LLM origin plausibility} (via its log loss), as depicted in Figure \ref{fig:dataset-pipeline}. The masking and filtering procedures are explained in detail below.
Google's Natural Questions (NQ) corpus from the Benchmark for Question Answering Research \parencite{NaturalQuestionsGoogleBenchmark} was used for evaluation purposes.

\iffalse
The HC3 dataset consists of >24k entries with questions regarding various topics, each answered by real humans and ChatGPT, respectively. Instead of reusing the existing GPT answerers, new ones using \textit{Qwen 1.5-4B} \parencite{qwen_2023} were created.
This step is crucial as the same model was later used to receive the log loss for permuted answers. These permutations were generated using the T5-3B \parencite[]{T5ParaphraseGenerator2020}, a model for generating paraphrases of English sentences trained on the Google PAWS dataset \parencite[]{PAWSParaphraseAdversaries2019}. For each question, the best-paraphrased answers were selected according to three criteria: \textit{Similarity}, \textit{Coherence}, and \textit{LLM origin plausibility (log loss)}. The masking and filtering procedures are explained separately. For evaluation, Google's \textit{Benchmark for Question Answering Research} was used \parencite{NaturalQuestionsGoogleBenchmark}, a
\textit{Natural Questions} (NQ) corpus that contains questions from real users.
\fi

\subsubsection{Sampling and Masking}

For each question in the HC3 dataset, a response was generated using the \textit{Qwen 1.5-4B} model. These responses were then processed in the following manner: First, the answer was split into sentences and then tokenized using named-entity recognition (NER) to identify entities such as names, places, or numbers. In order to maintain the core message of a sentence, these tokens were preserved from substitution. For the remaining tokens, with the exception of the final token in each sentence, all possible 2-tuples of token combinations were generated for masking purposes. Each sentence could include multiple masks, as long as the masked portion did not exceed 15\% of the sentence. From all possible combinations for each sentence, ten were randomly drawn. Note that this number can also be chosen higher but was kept small in order to reduce the computational effort later on. The masked tokens were then filled using the T5 model \parencite[3B;][]{T5ParaphraseGenerator2020} by generating ten paraphrased sentences per masked combination. Duplicates were discarded, ensuring variability in the output. This approach follows a methodology similar to the one described by \textcite{mitchell_detectgpt_2023}.

%For a sentence with $n$ non-entity tokens, the number of possible 2-tuple combinations is given by $m=\tfrac{n!}{2!(n-2)!}$, with $m\leq 0.15n$. 
%By generating ten permutations for each sample and sampling each permutation ten times using the T5 model, 2.43 million sentences are generated to generate 24,300 training samples. This is already a high computational effort, which explains why the numbers of permutations, samples, and the underlying model have been kept relatively small. 

\subsubsection{Filtering Criteria}\label{sec:filtering_exp3}

The filtering criteria were designed to select paraphrases that show the lowest likelihood of being generated by the corresponding large language model (LLM), consequently being difficult to detect as machine-generated. However, these paraphrases were also required to maintain a high degree of similarity to the original text in order to prevent semantic drift (i.e., a high similarity score is desirable). Furthermore, coherence was a crucial factor in ensuring that the paraphrases remained linguistically correct. Sentences with high similarity but incoherent or linguistically incorrect structures may evade detection by LLM classifiers due to unusual word choices. Yet, they are clearly distinguishable from human responses, making them unsuitable for the intended use. 
Paraphrases not meeting the mentioned criteria were discarded, while the best of the remaining ones were chosen.

\subparagraph{\textbf{Similarity}} The assessment of semantic similarity was conducted using the all-MiniLM-L6-v2 model \parencite{MiniLM}, which measures the cosine similarity between sentence vectors. This approach ensured that the selected paraphrases did not deviate excessively from the meaning of the original sentences, thus preserving the semantic content required for reliable training data.

\subparagraph{\textbf{Coherence}} Coherence or linguistic acceptability was evaluated through a DeBERTa classifier \parencite{DeBERTa}, trained on the \textit{Corpus of Linguistic Acceptability} dataset \parencite[CoLA;][]{CoLA}. This ensured that the paraphrased sentences retained grammatical correctness and avoided structural errors that might result in linguistically faulty outputs.

\subparagraph{\textbf{LLM origin plausibility}} The paraphrased sentences were evaluated using the model Qwen1.5-4B-Chat itself \parencite{qwen_2023}, to calculate the log loss. The paraphrase with the highest log loss, indicating the lowest likelihood of being generated by the model, was selected. 
This criterion aimed to identify paraphrases that were most likely to evade detection by the LLM classifier. As shown in Table \ref{tab:permutationTrainSet}, the sentence: \enquote{\textit{When you use Tor, your online activity is \textbf{encrypted and} can't be traced back to your \textbf{real identity}}} is paraphrased with the goal to identify the combination with the highest cross-entropy loss.

\begin{table}% [h]
\tabcolsep=0pt%
\TBL{\caption{Example of Trainset Generation. %:\newline .
\label{tab:permutationTrainSet}}}
{\begin{fntable}
\begin{tabular*}{\textwidth}{@{\extracolsep{\fill}}lr@{}}\toprule%
\TCH{Paraphrased sentence} & \TCH{Cross-entropy loss} \\\midrule
        {\fontfamily{cmss}\selectfont When [...] encrypted and [...] real location.} & $4.582982$\,\\
        {\fontfamily{cmss}\selectfont When [...] encrypted and [...] computer.} & $4.613022$\,\\
        {\fontfamily{cmss}\selectfont When [...] not visible to most websites and [...] true location.} & $4.570329$\,\\
        {\fontfamily{cmss}\selectfont When [...] encrypted so that it [...] ISP.} & $4.608329$\,\\
        {\fontfamily{cmss}\selectfont When [...] anonymous, and it [...] address.} & $4.626064$\,\\
        {\fontfamily{cmss}\selectfont When [...] secure and [...] actual computer.} & $4.691911$\,\\
        {\fontfamily{cmss}\selectfont When [...] secure and [...] true location.} & $4.651963$\,\\
        {\fontfamily{cmss}\selectfont When [...] anonymous and [...] actual home address or computer.} & $4.637465$\,\\
        {\fontfamily{cmss}\selectfont When [...] anonymous and [...] ISP.} & $4.672797$\,\\
\botrule
\end{tabular*}%
\footnotetext[]{\emph{Note:} Original sentence: When you use Tor, your online activity is \textbf{encrypted and} can't be traced back to your \textbf{real identity}.}
\end{fntable}}
\end{table}

\subsubsection{Filtering and Scoring}
Initially, all generated examples with a cosine similarity of less than 0.9 were discarded. 
For coherence or linguistic acceptability, two rules were applied. If the original sentence had a coherence score of 0.9 or higher, any paraphrased sentences scoring below this threshold were rejected. Likewise, if the original sentence had a coherence score below 0.9, any paraphrases with a delta greater than $\pm 0.05$ compared to the original were discarded, allowing only minor linguistic changes. This second delta rule is necessary because an original sentence may already have been written in a bad linguistic style. So, every paraphrased sentence would have been discarded. The remaining samples were then sorted by log loss, with the highest log loss sample being selected for each original sentence. This process yielded a training set comprising 24.300 paraphrases.

\subsubsection{Paraphrasing Model}\label{sec:model_exp3}
The paraphrasing model is based on the T5-11B transformer \parencite{2020t5}, trained for three epochs with a learning rate of $\alpha=10^{-4}$. During training, the prompt “paraphrase:~” was added to guide the model in generating paraphrases. T5 stands for Text-to-Text Transfer Transformer, and this is the reason why the model has been favored over other transformer models and architectures. It takes a task and the actual input as a combined input and produces the result as an output without additional explanation, introduction, hints, etc.
Therefore, the paraphrasing task is treated similarly to a translation task, with the difference that the model is taught to introduce minor changes to the original sentences.
Since the train data have been designed only to introduce small changes of about 15\% of the tokens, the model can easily be applied recursively without removing too much of the actual meaning while being applied. This allows for the degree of change that should be applied to the original text to be adjusted.

\subsubsection{Comparison Models}\label{sec:compmodels_exp3}
In order to compare the conducted approach to existing ones in the scientific field, two methods by \textcite{krishna_paraphrasing_2024} and \textcite{sadasivan_can_2024} are used. To make the comparison as significant as possible, the same data are used for \textcite{krishna_paraphrasing_2024}. This is not feasible for \textcite{sadasivan_can_2024} since their results have been human-rated, as explained below. Therefore, the results are converted for a better comparison. 

\paragraph{\textbf{Lexical diversity approach}}
Dipper by \textcite{krishna_paraphrasing_2024} uses a fine-tuned T5-11B transformer model. In a single paraphrasing step, the Dipper model takes two parameters, \textit{Lexical Diversity} and \textit{Order Diversity}. This means that instead of adjusting the degree of change by applying a paraphrasing model recursively, Dipper uses the parameter \textit{Lexical Diversity}. Furthermore, Dipper is designed as a general-purpose paraphrasing model and not fine-tuned to hide a specific model such as Qwen1.5-4b-chat \parencite{qwen_2023}.

\paragraph{\textbf{Paraphrasing approach}}
A comparable approach was conducted by \textcite{sadasivan_can_2024}, who also used paraphrasing. Instead of training their own models for the hiding of the machine-generated text, the authors combined several pre-trained language models. Although optimized regarding detectability, the authors also report measures regarding \textit{grammar or text quality} (similar to \textit{linguistic acceptability}) and \textit{content preservation} (comparable with \textit{cosine similarity}). Both were human-rated on a Likert scale, thus making them less comparable. 

%The mentioned approach also surpassed \citeauthor{sadasivan_can_2024} who used recursive paraphrasing, too, but combined several pre-trained paraphrasing models instead of fine-tuning them.

% \subsection{Evaluation}\label{sec:evaluation_exp3}
\subsection{Results}\label{sec:evaluation_exp3}

Without the application of any hiding model, the baseline detection rate was 88.9\% and the linguistic acceptability 74.1\%. Cosine similarity was, of course, 100\%.

After applying the here proposed trained hiding model, all values decreased. The detection rate went down quite fast and reached 8.7\% after 10 permutations. At the same time, both remaining variables were still quite high: Linguistic acceptability reached 65.5\% while remaining 82.6\% similarity to the original sentence. Both variables showed low variance across permutations, with mean values of $M=68.2\%$ ($SD=2.3\%$) regarding acceptability and $M=89.0\%$ ($SD=4.5\%$) for similarity.

Starting with the same baseline values, the comparing model of \textcite{krishna_paraphrasing_2024} leads to a similar linguistic acceptability of 72.5\%. The detection rate decreased to 15.4\%, and the cosine similarity also went down, reaching 38.9\%. Overall applications, mean acceptability was high with low variance $M=72.6\%$ ($SD=0.6\%$), contrarily to the similarity with $M=76.4\%$ ($SD=22.0\%$).

Hence, the detection rate is more than twice as high as with the proposed model, which did not lead to a significant decrease in similarity.

Results reported by \textcite{sadasivan_can_2024} are less comparable since a different dataset was used. Additionally, the evaluation was done by human raters on a Likert scale regarding \textit{grammar or text quality} (similar to \textit{linguistic acceptability}) and \textit{content preservation} (comparable with \textit{cosine similarity}). Rescaling their results leads to the following values after five permutations: Acceptability of 76.8\% and similarity of 67.5\% were both high, but the detection rate only decreased to 60.9\%. Without reporting the number of permutations, the minimal detection rate is at 58.1\%. 

% x variable value 

% 1     5 detect    60.9       
% 2     5 smlrty    67.5       
% 3     5 accept    76.8       

% # smlrty = Content preservation
% # accept = Grammar or text quality

%After ten permutation iterations over 1000 answers to the Google Natural Questions (NQ) dataset, the average similarity was still 0.83, while the detection rate of the 1000 samples dropped to 9\%. The application of this recursive paraphrasing model is illustrated in table~\ref{tab:exmpl_exp3}.

\newcommand{\smallstrut}{\rule[-0.3ex]{0pt}{1.5ex}}
\makeatletter
\newcommand\hglt[1]{\tikz[baseline]{\node[fill=blue!10, anchor=base, rounded corners=2pt] {\smallstrut #1};}}
\makeatother

\newcommand\Hglt[1]{\tikz[baseline]{\node[fill=blue!30, anchor=base, rounded corners=2pt] {\smallstrut #1};}}

\begin{table}%[t]
\tabcolsep=2pt%
\TBL{\caption{Example of original Answer vs. Paraphrases to the Question: Who was Michelangelo?\label{tab:exmpl_exp3}}}
{\begin{fntable}
\begin{tabularx}{\textwidth}{c X c}
\toprule
$i$ & Response & cos \\\midrule

        0 & {\fontfamily{cmss}\selectfont\small Michelangelo was a very famous painter and sculptor from a long time ago. He was born in Italy and he made many beautiful pictures and statues that people still love today. Some of his most famous works include the sculpture of David and the painting of the Sistine Chapel ceiling.} & $1.00$ \\

        1 & {\fontfamily{cmss}\selectfont\small Michelangelo was a very \hglt{talented} painter and sculptor \hglt{that lived} a long time ago. He was born in Italy and he made many \hglt{$\footnotesize\times$} pictures and statues \hglt{which are known and loved} today. Some of his \hglt{more} famous works include the sculpture of David and the painting of the Sistine Chapel ceiling.} & $0.98$ \\
        
        2 & {\fontfamily{cmss}\selectfont\small Michelangelo was a very \Hglt{talented} painter and sculptor \Hglt{that} \hglt{became popular} a long time ago. He was born in Italy\hglt{, where} he made many \Hglt{$\footnotesize\times$} pictures and statues \hglt{that} \Hglt{are} \hglt{famous} \Hglt{and loved} today. Some of his \Hglt{more} famous works include the sculpture of David and the painting of the Sistine Chapel.} & $0.97$ \\
        
        3 & {\fontfamily{cmss}\selectfont\small Michelangelo was a very \Hglt{talented} painter and sculptor \hglt{who lived} a long time ago. He was born in Italy, where he made many \Hglt{$\footnotesize\times$} \hglt{paintings} and statues that are \hglt{well known even} today. Some of his \hglt{best work includes} the sculpture of David and the painting of the Sistine Chapel.} & $0.95$ \\
        4 & {\fontfamily{cmss}\selectfont\small Michelangelo was a very \Hglt{talented} painter and sculptor \Hglt{who lived} \hglt{during the Renaissance.}. He was born in Italy\hglt{$\footnotesize\times$ and} he made many \Hglt{$\footnotesize\times$ paintings} and statues that are \hglt{also} \Hglt{well known even} today. Some of his \Hglt{best work includes} the sculpture of David and the \hglt{paintings in} the Sistine Chapel.} & $0.94$ \\

\botrule
\end{tabularx}
\footnotetext[]{{\textit{Note:}} Original promt: {\fontfamily{cmss}\selectfont 'Who was Michelangelo? Explain to me like a child'}. The original answer given by Qwen1.5-4B-Chat is marked as $i=0$ and recursive iterations as $ I \geq 1$, exemplary for the first four out of ten. Colorization reflects qualitative change across applications: Light color highlights changes to the previous response, and dark color represents overtaken changes (i.e., changes compared to the original response). Deleted words or punctuation are xed out. Similarity is quantized as cosine similarly (cos).}
\end{fntable}}
\end{table}

As the number of paraphrasing iterations increases, the probability of the sentence being identified as machine-generated (i.e. classifier label = 0) decreases significantly. At the same time, the likelihood of it being linguistically acceptable remains relatively stable. Furthermore, the degree of semantic similarity also decreases only marginally.

\begin{figure}[t] %[ht]
    {\caption{Results of the proposed Model vs. Reference Models}
    \label{fig:results3}}
    \FIG{\includegraphics[width=0.99\textwidth]{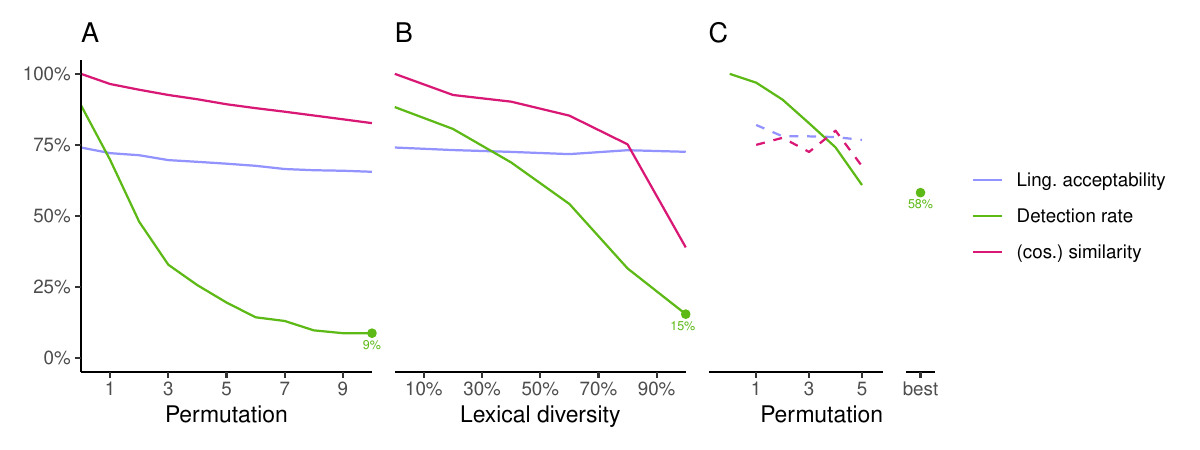}}
    \flushleft
    \footnotesize
    \emph{Note:} Comparison of the results achieved with \textbf{\sffamily A}. the proposed model in comparison to \textbf{\sffamily B}. the \textit{Discourse Paraphraser} DIPPER by \textcite{krishna_paraphrasing_2024} and \textbf{\sffamily C}. the paraphrasing model by \textcite{sadasivan_can_2024}. \citeauthor{sadasivan_can_2024} report values for \textit{grammar or text quality} (comparable to \textit{linguistic acceptability}) and \textit{content preservation} (matched to \textit{similarity}), both manually labeled on a Likert scale of 1 to 5 (scaled here for better comparability but marked by dashed line representation). Those values are only provided for permutations 1--5 with detection rates, i.e., it is unclear at which permutation their 'best' result occurred. 
\end{figure}

%A comparative analysis was conducted against the Dipper approach proposed by \textcite{krishna_paraphrasing_2024}, employing a fine-tuned T5-11B transformer model. However, there are significant differences between both models. In a single paraphrasing step, the Dipper model takes two parameters, \textit{Lexical Diversity} and \textit{Order Diversity}. This means that instead of adjusting the degree of change by applying a paraphrasing model recursively, Dipper uses the parameter \textit{Lexical Diversity}. Furthermore, Dipper is designed as a general-purpose paraphrasing model, adaptable to any language model, while our approach targets a specific model, namely Qwen1.5-4b-chat \parencite{qwen_2023}.

Both methods' performance was compared by using ten iterations of the perturbation model and a lexical diversity range of 0\% to 100\% for the Dipper model. The benchmarks were performed on the 1,000 questions sourced from Google's NQ corpus \parencite{NaturalQuestionsGoogleBenchmark}. After six iterations of recursive paraphrasing, the detection rate dropped from 88.9\% to 14.3\%, while maintaining a high cosine similarity of 88\% (cf. Fig.~\ref{fig:results3}). 
In contrast, achieving a comparable result with the Dipper model required setting a lexical diversity of 100\%, which reduced the cosine similarity to 39\%. These results demonstrate that fine-tuning a paraphrasing model to evade detection for a specific generative model can achieve superior performance compared to a general-purpose paraphrasing model.

Summarized, the main advantages of the conducted new approach compared to \textcite{krishna_paraphrasing_2024} and \textcite{sadasivan_can_2024} are a lower detectability while preserving a higher cosine similarity as well as a constant linguistic acceptability.

\subsection{Discussion}
The evaluation was conducted using Google's Natural Questions corpus \parencite[NQ;][]{NaturalQuestionsGoogleBenchmark}, as described in the related work. 
In contrast to the evaluation method proposed by \textcite{mitchell_detectgpt_2023}, this study employs a perturbation technique that modifies the responses of a language model with the objective of increasing log loss. In particular, a paraphrasing model based on the T5-11B transformer is employed to generate sentences with increased log loss. This approach enables the generation of paraphrases that progressively deviate from the original sentence, increasing its distance from the language model's typical output. Despite these modifications, the linguistic acceptability of the generated sentences is preserved.

Experiment 3 proved that post-generation paraphrasing helps to avoid detection. This is particularly useful for bigger LLMs that are harder to fine-tune or for those that are inaccessible to the user. Although paraphrasing approaches have already been proven to be applicable \parencite{sadasivan_can_2024, krishna_paraphrasing_2024}, it is demonstrated that the evasion results are even better if the paraphrasing model is tailored to hide one specific model. The results outperformed those of \citeauthor{krishna_paraphrasing_2024} by adding the recursive paraphrasing feature. The mentioned approach also surpassed \citeauthor{sadasivan_can_2024}, who used recursive paraphrasing, too, but combined several pre-trained paraphrasing models instead of fine-tuning them. Nevertheless, this comparison also has its limitations since \citeauthor{krishna_paraphrasing_2024} used a parameter for lexical diversity instead of recursively applying the model. Likewise, \citeauthor{sadasivan_can_2024} used a Likert scale instead of a cosine similarity of sentence vectors to compare the permutations to the original output.
 
\section{Conclusions and Outlook}

We discussed the improvement of LLMs in terms of reasoning performance, which will further complicate LLM detection in the future. In the literature review, it is noted that LLMs are no longer recognizable to humans. This makes machine-based LLM detection vital. In the experimental series, it is shown that all types of state-of-the-art classifiers can be circumvented with sufficient effort. In particular, shallow detectors (\hyperref[sec:exp1]{Exp. 1}), transformer-based detectors (\hyperref[sec:exp2]{Exp. 2}) and zero-shot-based detectors (\hyperref[sec:exp3]{Exp. 3}) were attacked succesfully. For simple detectors, a small adjustment of the generative LLMs proved to be sufficient, such as adjusting the temperature (parameter $\tau$ in \hyperref[sec:exp1]{Exp. 1}). However, more sophisticated detectors, as used in practice, require a better, more elaborate adjustment of the generative LLMs. During reinforcement learning, these models tend to acquire strategies that make them unrecognizable to detectors (e.g., inserting strings of special characters). While increasing evasion, these techniques tend to impede both the syntactical and semantical quality of the altered text.
%However, the resulting output loses quality as it becomes incorrect grammatically and/or semantically. 
To counteract these model tendencies, two options were presented. Firstly, such behaviors were penalized during reinforcement learning by introducing additional constraints (cf. \hyperref[sec:exp2]{Exp. 2}). Secondly, paraphrasing was employed, where the algorithm filtered results based on the highest similarity to the original answer, the best linguistic acceptance, and the lowest likelihood of being generated by a language model (i.e., the lowest log loss, cf. \hyperref[sec:exp3]{Exp. 3}). With an evasion rate of $>90\%$, the model trained on the resulting dataset performed not only better than comparable models, but its results were also most similar to the original response.

In the experimental series, the language used on social media, including X (formerly Twitter), was examined, and the results were generalized to internet language in general. Hence, for any current written indirect communication on the web, it is currently impossible to tell whether it is coming from a human or a machine (LLM) if the author took actions to hide the origin of the text. 
This everlasting battle between creating new detectors and crafting new evasion techniques will continue.
%It is evident that researchers will create new detectors and we or colleagues will attack and evade these too. 
Apart from the latest technical developments in this cat-and-mouse game, the question arises which implications non-detectable LLM communication will have. Of course, there are a lot of potential as well as danger. Among these is a systematic change of opinion, the spread of misinformation, or even fake news. It has been shown that people adapt their opinions to the majority, as can be seen, for example, by \textcite{asch2016effects, asch1956studies} in his still relevant conformity experiments. Although an apparently false opinion was propagated in these experiments, around a third of the test subjects adapted their own opinion to that of the other test subjects (actors) versus less than 1\% without conformity pressure (control). In this regard, a transfer to social media is highly plausible: Users should hypothetically adapt their opinions to a majority of bots (posting LLM-based content) as long as those bots share the same opinion uniformly. Further research is needed for concrete proof. However, if conformity effects can be transferred to social media, it might be possible to influence the beliefs of citizens consequently. This could have an extreme social impact, such as influencing elections. 

In contrast, a range of advantages exists, resulting from LLMs. For example, the replacement of low-level work, such as answering frequently asked questions repetitively. However, this also bears risks. For example, \textcite{shumailovCurseRecursionTraining2024} measured the keystrokes of the agents behind the Mechanical Turk. Too few keystrokes for too long texts indicate that they must have copied their answers from ChatGPT or comparable applications instead of writing them themselves. 
The problem behind this practice lies in the quality of the resulting answers. When answers are designed to be used for model training but originate from LLMs in the first place, the performance of new LLMs decreases. In the long run, feeding models with model output itself leads to systematic model destruction \parencite{shumailov2024ai}. This is a real problem that might become even more relevant in the future, especially if there is no way to classify input data as human originated reliably. Already now, over 1\% of academic studies are said to be generated by LLMs or with its help \parencite{gray_chatgpt_2024}. This makes the affected papers useless for long-term model training. Besides the problem for future model training, Gray also denotes a potential effect on human behavior: The pure consumption of such texts might not only influence \textit{what} we think in terms of facts, but also \textit{how} we think to write correctly. That way, LLM text consumption might influence personal writing habits.

To support informed decision-making in applications where large language models contribute to hybrid outputs, future research should prioritize the development and implementation of tamper-resistant watermarking techniques that are compatible with diverse text generation methods. These watermarking strategies allow for a clear identification of machine-generated texts while preserving the semantic integrity of the content, but are yet to be deployed more commonly. Existing approaches range from lexical and syntactical permutations over logit-based generation patterns to watermarks anchored in the LLM training procedure \parencite{DBLP:journals/corr/abs-2409-00089,liu2024survey}. From an ethical perspective, the adoption of watermarking aligns with principles of transparency, accountability, and trustworthiness. 

The widespread adoption of undetectable large language models introduces moral challenges, such as the risks of manipulation of public discourse and the dissemination of biased or wrong narratives. Addressing these challenges requires a comprehensive approach to mitigation. Besides enhancing detectability via watermarking techniques, developing regulatory frameworks tailored to the ethical use of LLMs should be promoted. Further strategies may include promoting collaboration between developers, policymakers, and ethicists to establish industry-wide standards and guidelines for responsible LLM development. Finally, supporting campaigns to foster critical thinking skills among users to better evaluate the credibility of information, especially on high-volume platforms such as social media.

To conclude, large language models and their generated content bring huge opportunities to the table, but this also makes their detection more and more difficult, up to the point that they become undetectable. Society must know this risk, try to mitigate it as well as possible, and find coping strategies in case detection becomes impossible.

\newpage

% \begin{appendix}
% \section{Appendix. Title for Appendix Section}\label{appendixA}
% Appendix text here.
% \end{appendix}

\begin{Backmatter}

\paragraph{Acknowledgments}

The authors would like to thank the \textit{System Sciences Chair for Communication Systems and Network Security}, as well as the audience of the \textit{57th Hawaii International Conference} for their valuable discussions and feedback regarding the first version of this work \parencite[see][]{schneider2024HowWell}. Special thanks also go to the CODE Research Institute for providing the hardware.

\paragraph{Funding Statement}
The authors acknowledge the financial support by the Federal Ministry of Education and Research of Germany in the program of “Souverän. Digital. Vernetzt.”. Joint project 6G-life, project identification number: 16KISK002

\paragraph{Competing Interests}
None. 

\paragraph{Data Availability Statement}
\textbf{\textit{Datasets:}} The \textit{Human ChatGPT Comparison Corpus} \parencite[HC3;][]{ChatGPTvsHuman} is available at \href{https://huggingface.co/datasets/Hello-SimpleAI/HC3}{Hugging Face}: \url{https://huggingface.co/datasets/Hello-SimpleAI/HC3}. The \textit{Natural Questions Corpus}  \parencite[NQ;][] {NaturalQuestionsGoogleBenchmark} of \textit{Googles Benchmark for Question Answering Research} that was used for evaluation is also available at \href{https://huggingface.co/datasets/google-research-datasets/natural_questions}{Hugging Face}: \url{https://huggingface.co/datasets/google-research-datasets/natural_questions}. 
A script for generating the CNN/Daily Mail dataset \parencite{cnnDailyMailDataset} is provided at \href{https://github.com/google-deepmind/rc-data}{GitHub}: \url{https://github.com/google-deepmind/rc-data}. \textit{Note} that the repository contains the script only and the CNN and Daily Mail articles itself have to be downloaded separately using the Wayback Machine. A processed data set is provided by the \href{https://cs.nyu.edu/~kcho/DMQA/}{NYU}: \url{https://cs.nyu.edu/~kcho/DMQA/}. The Twitter data set can not be shared because of Twitters restrictions regarding their privacy policies. \textbf{\textit{Models:}} The used paraphrase generator with T5 \parencite[]{T5ParaphraseGenerator2020} is available at \href{https://zenodo.org/records/10731518}{Zenodo}: \url{https://zenodo.org/records/10731518}. The fine-tuning T5-XXL \textit{Discourse Paraphraser} \parencite[DIPPER;][]{krishna_paraphrasing_2024} can be downloaded at \href{https://huggingface.co/kalpeshk2011/dipper-paraphraser-xxl}{Hugging Face}: \url{https://huggingface.co/kalpeshk2011/dipper-paraphraser-xxl}. The pretrained language model \textit{Qwen 1.5-4B} \parencite{qwen_2023} can also be downloaded at \href{https://huggingface.co/Qwen/Qwen1.5-4B}{Hugging Face}: \url{https://huggingface.co/Qwen/Qwen1.5-4B}.

\paragraph{Environment for Replication Purposes}
All experiments were realized using a workstation equipped with four Nvidia A6000 Ada generation GPUs (48 GB VRAM each) and 512 GB of RAM. % The number of training samples for a BERT classifier was changed from 10.000 (\hyperref[sec:exp1]{Exp. 1}) to 100.000 (\hyperref[sec:exp2]{Exp. 2}) compared to the BoW classifiers.
For replication purposes, %of the experiments with the data stated above,
a lower-level machine (e.g., only one A6000 GPU) is sufficient but requires more data generation and training time. 
Especially for replication of the last experiment (\hyperref[sec:exp3]{Exp. 3}), it is strongly recommended to use the same level machine. Otherwise, a smaller model had to be chosen instead of the T5 XXL version.
%We also used Microsoft's \textit{DeepSpeed} library to train the T5 XXL model on multiple GPUs simultaneously.
To train the T5 XXL model on multiple GPUs simultaneously, Microsoft's \textit{DeepSpeed} library was used. 

\paragraph{Ethical Standards}
The research meets all ethical guidelines, including adherence to the legal requirements of the study country.

\paragraph{Author Contributions}
Funding acquisition: G.D.R.;
Project administration: S.S.;
Conceptualization: S.S.; 
Investigation: S.S., F.S., J.A.G.S.;
Methodology: S.S.; 
Data curation \& formal analysis: S.S.;
Visualization: J.A.G.S.;
Writing -- original draft: S.S., F.S., J.A.G.S.;
Writing -- review \& editing: S.S., F.S., J.A.G.S., G.D.R.;
Supervision: G.D.R.;
All authors approved the final submitted draft.

% \renewcommand\bibpreamble{By default, this template uses \texttt{bibtex} and adopts the AMS referencing style. However, the journal you’re submitting to may require a different reference style; specify the journal you're using with the class' \texttt{journal} option --- see lines 1--19 of \emph{sample.tex} for a list of options and instructions for selecting the journal.}

% If using any of the following journal options:
%   wet, dap, dce, eds, prm, flw, jdm, psy, rsm
% then use the \printbibliography line instead of:
% \bibliography{example}
\printbibliography

\end{Backmatter}

\end{document}